\documentclass{article}

\usepackage{microtype}
\usepackage{graphicx}
\usepackage{subfigure}
\usepackage{booktabs} 
\usepackage{algorithm}
\usepackage{algpseudocode}
\algblockdefx{MRepeat}{EndRepeat}{\textbf{repeat}}{}
\algnotext{EndRepeat}

\usepackage{hyperref}
\usepackage{math-common}

\usepackage[accepted]{icml2023}

\usepackage{amsmath}
\usepackage{amssymb}
\usepackage{mathtools}
\usepackage{amsthm}

\usepackage[capitalize,noabbrev]{cleveref}

\theoremstyle{plain}
\newtheorem{theorem}{Theorem}[section]
\newtheorem{proposition}[theorem]{Proposition}
\newtheorem{lemma}[theorem]{Lemma}

\theoremstyle{definition}

\theoremstyle{remark}

\usepackage[textsize=tiny]{todonotes}

\icmltitlerunning{Probabilistic Unrolling: Scalable, Inverse-Free Maximum Likelihood Estimation for Latent Gaussian Models}

\begin{document}

\twocolumn[
\icmltitle{Probabilistic Unrolling:
Scalable, Inverse-Free Maximum \\ Likelihood Estimation for Latent Gaussian Models}



\icmlsetsymbol{equal}{*}

\begin{icmlauthorlist}
\icmlauthor{Alexander Lin}{harvard}
\icmlauthor{Bahareh Tolooshams}{harvard}
\icmlauthor{Yves Atchad\'e}{bu}
\icmlauthor{Demba Ba}{harvard}
\end{icmlauthorlist}

\icmlaffiliation{harvard}{School of Engineering and Applied Sciences, Harvard University, Boston, MA, USA}
\icmlaffiliation{bu}{Department of Mathematics and Statistics, Boston University, Boston, MA, USA}

\icmlcorrespondingauthor{Alexander Lin}{alin@seas.harvard.edu}

\icmlkeywords{Machine Learning, ICML}

\vskip 0.3in
]



\printAffiliationsAndNotice{}  

\begin{abstract}
Latent Gaussian models have a rich history in statistics and machine learning, with applications ranging from factor analysis to compressed sensing to time series analysis. The classical method for maximizing the likelihood of these models is the expectation-maximization (EM) algorithm. For problems with high-dimensional latent variables and large datasets, 
EM scales poorly because it needs to invert as many large covariance matrices as the number of data points.  We introduce \emph{probabilistic unrolling}, a method that combines Monte Carlo sampling with iterative linear solvers to circumvent matrix inversion. Our theoretical analyses reveal that unrolling and backpropagation through the iterations of the solver can accelerate gradient estimation for maximum likelihood estimation.  In experiments on simulated and real data, we demonstrate that probabilistic unrolling learns latent Gaussian models up to an order of magnitude faster than gradient EM, with minimal losses in model performance.
\end{abstract}

\section{Introduction}
Latent variable models with Gaussian prior and Gaussian likelihood, i.e. \emph{latent Gaussian models} (LGMs), are popular and powerful tools within statistics and machine learning.  They have found applications in many settings, such as 
 factor analysis \citep{basilevsky2009statistical}, sparse Bayesian learning \citep{tipping2001sparse}, state-space models \citep{durbin2012time}, and neural linear models \citep{ober2019benchmarking}. In these models, the means and/or covariances of the Gaussian distributions are functions of parameters that must be optimized to fit observed data.  
 
The expectation-maximization (EM) algorithm \citep{dempster1977maximum} is a popular way to optimize the parameters by maximum likelihood estimation. One variant called \emph{gradient EM} \citep{lange1995gradient} implements the M-step through a single iteration of gradient descent. 
For problems with high-dimensional latent variables and many training examples, gradient EM scales poorly due to the need to invert as many large covariance matrices as the number of examples.

Advances in numerical linear algebra have demonstrated, in various contexts, that iterative solvers often provide a much faster alternative to matrix inversion \citep{saad2003iterative, ubaru2017fast, gardner2018gpytorch, lin2022covariance}. A separate, burgeoning literature on unrolled optimization has shown theoretical and practical benefits to differentiating through the iterations of deterministic optimizers \citep{maclaurin2015gradient, shaban2019truncated, ablin2020super, tolooshams2022stable, malezieux2021understanding}. This literature begs questions as to the potential benefits, in a latent variable setting, of unrolling the iterations of a sampler (i.e. stochastic solver), and differentiating through them.


\noindent \textbf{Contributions} We introduce \emph{probabilistic unrolling}, a computational framework that accelerates maximum likelihood estimation for large-scale, high-dimensional LGMs. 
 Our method provides a way to run gradient EM without matrix inversions.  Specifically, we design iterative linear solvers to yield the probabilistic quantities needed by the EM algorithm (i.e. posterior means and covariance samples).  Our method reduces the complexity of gradient EM from a cubic function of the latent dimension to a quadratic function in the general case, and a linear function in special cases.  

 We theoretically analyze the faithfulness of probabilistic unrolling to gradient EM when encountering two sources of error: (a) the \emph{statistical error} from using a finite number of covariance samples, and (b) the \emph{optimization error} from stopping the solver before convergence.  We provide bounds for both of these factors, producing insights on how to pick the number of samples and the number of solver iterations.  Finally, we show that our method can further improve its approximation to the true EM gradient by backpropagating through the unrolled iterations of the solver.

Probabilistic unrolling can be viewed as training a recurrent network in which each layer applies a matrix operation from the unrolled linear solver. 
We implement this highly structured architecture in modern deep learning frameworks to further benefit from GPU acceleration. We perform several experiments with simulated and real data, showing that probabilistic unrolling can fit LGMs of practical interest up to $70$ times faster than gradient EM.  Our code is available at \url{https://github.com/al5250/prob-unroll}.      

\section{Background: Latent Gaussian Model}
Let $\{\bidx y n\}_{n=1}^N$ denote $N$ i.i.d. observations, each associated with a latent variable $\bidx z n$. In a LGM, the \emph{prior} on each latent variable and \emph{likelihood} (i.e. conditional distribution) of each observation both follow Gaussian distributions,
\begin{align}
\bidx z n | \bd \theta&\sim \mathcal{N}(\bd \nu_{\bd \theta}, \bd \Gamma_{\bd \theta}^{-1}), \label{lgm} \\
\bidx y n| \bidx z n, \bd \theta &\sim \mathcal{N}(\bd \Phi_{\bd \theta} \bidx z n + \bd \eta_{\bd \theta}, \bd \Psi_{\bd \theta}^{-1}),  \quad n = 1, \ldots, N.\nonumber
\end{align}
The prior and likelihood depend on a set of \emph{canonical parameters} ($\bd \nu_{\bd \theta} \in \R^D, \bd \Gamma_{\bd \theta} \in \R^{D \times D}, \bd \Phi_{\bd \theta} \in \R^{M \times D}, \bd \eta_{\bd \theta} \in \R^M$, and a diagonal matrix $\bd \Psi_{\bd \theta} \in \R^{M \times M}$) that form the means and covariances of the Gaussian distributions.  The canonical parameters are themselves functions of the model's \emph{free parameters} $\bd \theta$, which are individual values that can be learned through maximum likelihood estimation. 


\textbf{Examples}
The LGM \eqref{lgm} generalizes many models within statistics and machine learning.  Some famous examples include 
(a) \emph{factor analysis}, a probabilistic generalization of PCA \citep{basilevsky2009statistical}, (b) \emph{sparse Bayesian learning}, a Bayesian approach to compressed sensing \citep{wipf2004sparse}, and (c) \emph{state-space models}, one of the most popular class of probabilistic time series models \citep{durbin2012time}. With the advent of deep learning, the LGM class has broadened to include complex, non-linear structures such as (d) \emph{neural linear models}, i.e. neural networks whose trainable weights correspond to free parameters \cite{ober2019benchmarking}. For each of these models (and others), we work out the definition of free parameters $\bd \theta$ and how they map to the canonical parameters in Appendix \ref{app:lgm-ex}.

\textbf{Missing Data}  In many applications of LGMs, $\bidx y n$ may have missing values, i.e. we may not observe all its entries.  To account for missing data, we assume that for each $n$, we observe $\bidx \ty n = \bidx \Omega n \bidx y n$, 
where the mask $\bidx \Omega n \in \R^{M_n \times M}$ is a row-wise subset of the $M \times M$ identity matrix.

\textbf{EM Inference} To fit the parameters $\bd \theta \in \Theta$ to data $\bidx \ty 1, \ldots, \bidx \ty N$, we perform maximum likelihood estimation or, equivalently, minimize the negative log-likelihood,
\begin{align}
\mathcal{L}(\bd \theta) :=& \frac{1}{N} \sum_{n=1}^N -\log p(\bidx \ty n| \bd \theta)  \label{nll} \\
=& \frac{1}{N} \sum_{n=1}^N -\log \int p(\bidx \ty n | \bidx z n, \bd \theta) p(\bidx z n | \bd \theta) d \bidx z n. \nonumber
\end{align}
Due to the latent variable $\bidx z n$, one common approach to minimizing \eqref{nll} is to use the \emph{expectation-maximization} (EM) algorithm \citep{dempster1977maximum}.  EM revolves around the $\mathcal{Q}$-function, which is defined for any $\{\bd \theta_1, \bd \theta_2\} \in \Theta \times \Theta$ as
\begin{align}
\mathcal{Q}(\bd \theta_1 | \bd \theta_2) &:= \frac{1}{N} \sum_{n=1}^N \idx q n (\bd \theta_1 | \bd \theta_2) \label{e-step}, \\
\idx q n (\bd \theta_1 | \bd \theta_2) &:= \mathbb{E}_{p(\bidx z n | \bidx \ty n, \bd \theta_2)}[- \log p(\bidx z n, \bidx \ty n|  \bd \theta_1)]. \nonumber 
\end{align}
The $\mathcal{Q}$-function is called the \emph{expected complete-data negative log-likelihood} because it averages the negative log-likelihood of the observed data $\bidx y n$ and the unobserved data $\bidx z n$ over all possible realizations of $\bidx z n$ 
\citep[Ch. 9]{bishop2006pattern}.  EM iterations repeatedly alternate between constructing $\mathcal{Q}$ and minimizing it to make progress on $\mathcal{L}$: Given a current solution $\bd \theta^\text{old}$, the \emph{E}-step computes the posterior distribution $p(\bidx z n | \bidx \ty n, \bd \theta^\text{old})$ to form the function $\mathcal{Q}(\bd \theta | \bd \theta^\text{old})$, defined for all $\bd \theta \in \Theta$.   The \emph{M}-step then finds a new solution $\bd \theta^\text{new}$ such that $\mathcal{Q}(\bd \theta^\text{new} | \bd \theta^\text{old}) \leq \mathcal{Q}(\bd{\theta}^\text{old} | \bd \theta^\text{old})$.  This guarantees that $\mathcal{L}(\bd \theta^\text{new}) \leq \mathcal{L}(\bd \theta^\text{old})$.  

Variants of EM differ in how they implement the \emph{M}-step. \emph{Classical EM} \citep{dempster1977maximum} solves an optimization problem, i.e. $\bd \theta^\text{new} := \arg \min_{\bd \theta \in \Theta} \mathcal{Q}(\bd \theta | \bd \theta^\text{old})$.  
We focus on a computationally-simpler alternative called \emph{gradient EM}  \citep{lange1995gradient, balakrishnan2017statistical}, 
\begin{align}
\bd \theta^\text{new} &:= \bd \theta^\text{old} - \alpha \cdot \nabla_1 \mathcal{Q}(\bd \theta^\text{old} | \bd \theta^\text{old}), \label{grad-em} 
\end{align}
where $\alpha \in \R$ is the step size and $\nabla_1 \mathcal{Q}$ means the gradient with respect to the first argument of $\mathcal{Q}$, as defined in \eqref{e-step}.  


 \textbf{EM for the LGM} For latent Gaussian models, $\mathcal{Q}$ and its gradient are computable in closed-form.  Each $\idx q n$ in \eqref{e-step} simplifies to (dropping the index $n$ for convenience):
\begin{align}
q(\bd \theta_1 | \bd \theta_2) = \frac{1}{2} \bd \mu_{\bd \theta_2}^\top \bold A_{\bd \theta_1} \bd \mu_{\bd \theta_2} &- \bd b_{\bd \theta_1}^\top \bd \mu_{\bd \theta_2} \label{qn} \\
&+ \frac{1}{2} \text{Tr}(\bold A_{\bd \theta_1} \bold \Sigma_{\bd \theta_2}) + c_{\bd \theta_1}, \nonumber 
\end{align}
where, for all $\bd\theta \in \Theta$, we define the quantities
\begin{align}
\bold A_{\bd \theta} &:= \bd \Gamma_{\bd \theta} + \bd \Phi_{\bd \theta}^\top \bd \Omega^\top \bd \Omega \bd \Psi_{\bd \theta} \bd \Omega^\top \bd \Omega \bd \Phi_{\bd \theta}, \label{def-a}\\
\bd b_{\bd \theta} &:= \bd \Gamma_{\bd \theta}  \bd \nu_{\bd \theta}  + \bd \Phi_{\bd \theta}^\top \bd \Omega^\top \bd \Omega  \bd \Psi_{\bd \theta} \bd \Omega^\top (\bd \ty - \bd \Omega \bd \eta_{\bd \theta}), \nonumber \\
c_{\bd \theta} &:= \tfrac{1}{2} (\bd \ty -\bd \Omega \bd \eta_{\bd \theta})^\top \bd \Omega \bd \Psi_{\bd \theta} \bd \Omega^\top (\bd \ty -\bd \Omega \bd \eta_{\bd \theta}) + \tfrac{1}{2}\bd \nu_{\bd \theta}^\top \bd \Gamma_{\bd \theta} \bd \nu_{\bd \theta} \nonumber \\
& \quad \quad \quad -\tfrac{1}{2} \log \det \bold \Omega \bold \Psi_{\bd \theta} \bold \Omega^\top  -  \tfrac{1}{2} \log \det \bd \Gamma_{\bd \theta}, \nonumber
\end{align}
and the posterior $p(\bd z | \bd \ty, \bd \theta) \sim \mathcal{N}(\bd \mu_{\bd \theta}, \bd \Sigma_{\bd \theta})$  is given by
\begin{align}
\bd \mu_{\bd \theta} := \bd \Sigma_{\bd \theta}  \bd b_{\bd \theta}, & & \bd \Sigma_{\bd \theta} := \bold A_{\bd \theta}^{-1}. \label{mvn}
\end{align}
The derivation for equations \eqref{qn}-\eqref{mvn} is given in Appendix \ref{qstar-deriv}.

\textbf{Computational Challenges} Gradient EM involves computing the gradient of \eqref{qn} (which we call the \emph{exact gradient}),
\begin{align}
\bd g^\star(\bd \theta) := \nabla_1 q(\bd \theta | \bd \theta). \label{exact-grad}
\end{align}
Since $\bd g^\star(\bd \theta)$ depends on the posterior moments $(\bd \mu_{\bd \theta}, \bd \Sigma_{\bd \theta})$, it requires inverting a large matrix of size $D \times D$. This has time cost $\mathcal{O}(D^3)$ and storage cost $\mathcal{O}(D^2)$, which becomes prohibitive for large $D$. Furthermore, for $N$ different data vectors, we need to compute $N$ posterior moments $(\bidx[\bd \theta] \mu n, \bidx[\bd \theta] \Sigma n)$, which requires $N$ separate matrix inversions.  

We now arrive at the main goal of the paper: In the ensuing sections, we introduce a computational framework called \emph{probabilistic unrolling} that can provably accelerate gradient EM by avoiding explicit matrix inversions. This allows us to fit latent Gaussian models at substantially greater scale in high dimensions $D$ and for large dataset sizes $N$. 

\section{Method: Probabilistic Unrolling}
\label{sec:bunrl}

Probabilistic unrolling circumvents matrix inversion by iteratively solving multiple linear systems in parallel. We design the systems to perform posterior inference, i.e. the solutions are the posterior mean $\bd \mu_{\bd \theta}$ and covariance samples distributed as $\mathcal{N}(\bd 0, \bd \Sigma_{\bd \theta})$.  We use these quantities to estimate the EM objective \eqref{qn} and its gradient $\bd g^\star(\bd \theta)$ \eqref{exact-grad}. This process requires less time and memory than computing $\bd g^\star(\bd \theta)$ directly. We also show that backpropagating through the linear solvers further improves our estimation of $\bd g^\star(\bd \theta)$.

\begin{figure}[ht]
\vspace{-0.5em}
\begin{center}
\centerline{\includegraphics[scale=0.23]{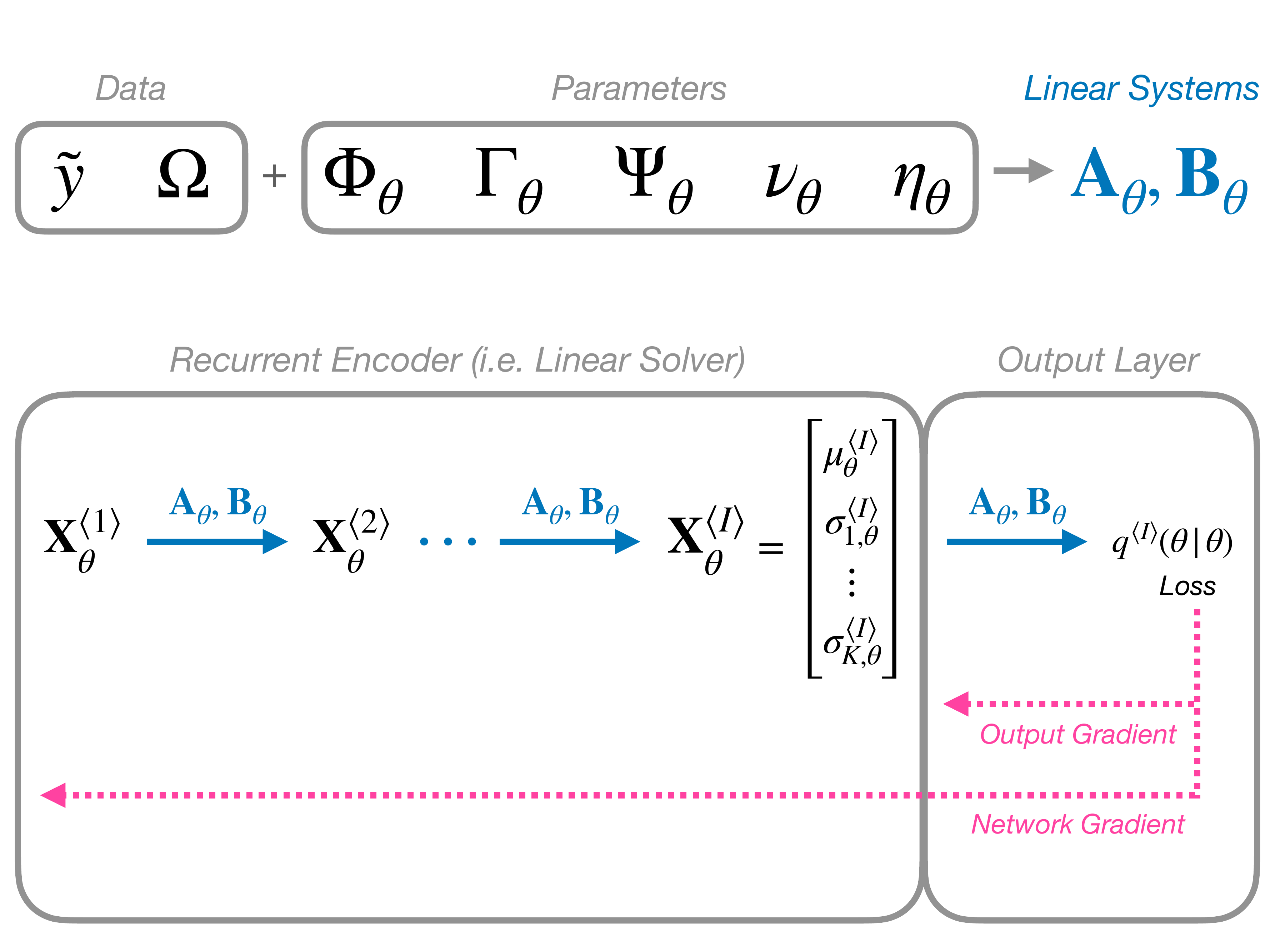}}
\caption{The probabilistic unrolling architecture: The data point $\btilde y$, mask $\bd \Omega$, and parameters $\bd \theta$ define the linear operator $\bold A_{\bd \theta}$ and construct the matrix $\bold B_{\bd \theta}$.  A linear solver, unrolled for $I$ steps, solves the matrix equation $\bold A_{\bd \theta} \bold X_{\bd \theta} = \bold B_{\bd \theta}$, yielding the posterior mean $\bd \mu_{\bd \theta}$ and samples $\{\bd \sigma_{1, \bd \theta}\}_{k=1}^K$ with covariance $\bd \Sigma_{\bd \theta}$.  These posterior quantities are used to compute either the output gradient \eqref{out-grad} or network gradient \eqref{net-grad} to approximate the true EM gradient.}
\label{arch}
\end{center}
\vspace{-2em}
\end{figure}

From a deep learning perspective, the overall method looks like a recurrent network (Fig. \ref{arch}). We can view the unrolled sequence of solver iterations as a \emph{recurrent encoder}, with weights $\bd \theta$, that takes the observed data $\bd \ty$ and refines \emph{hidden states} representing the distribution $p(\bd z | \bd \ty, \bd \theta)$. The hidden states are then passed through an \emph{output layer}, also parameterized by $\bd \theta$, to evaluate the loss \eqref{qn}. Training this network is equivalent to running gradient EM for the LGM. 

\subsection{Monte Carlo Gradient EM}

In high-dimensional settings, inverting a matrix to compute $\bd \Sigma_{\bd \theta}$ is the main bottleneck of \eqref{qn}. The first step of our method replaces the trace term containing $\bd \Sigma_{\bd \theta}$ with an unbiased estimator.  Given any square matrix $\bold A$ and a sample $\bd \sigma_{\bd \theta} \sim \mathcal{N}(\bd 0, \bd \Sigma_{\bd \theta})$, it follows that $\E[\bd \sigma_{\bd \theta}^\top \bold A \bd \sigma_{\bd \theta}] = \text{Tr}(\bold A \bd \Sigma_{\bd \theta})$ \citep{skilling1989eigenvalues, hutchinson1989stochastic}.  Using $K > 1$ independent samples $\bd \sigma_{1, \bd \theta}, \ldots, \bd \sigma_{K, \bd \theta} \sim \mathcal{N}(\bd 0, \bd \Sigma_{\bd \theta})$ (to reduce variance) leads to the following approximation of \eqref{qn},
\begin{align}
q^\#(\bd \theta_1 | \bd \theta_2) := &\frac{1}{2} \bd \mu_{\bd \theta_2}^\top \bold A_{\bd \theta_1} \bd \mu_{\bd \theta_2} - \bd b_{\bd \theta_1}^\top \bd \mu_{\bd \theta_2}\label{mc-qn} \\
&\quad \quad + \frac{1}{2K} \sum_{k=1}^K \bd \sigma_{k, \bd \theta_2}^\top \bold A_{\bd \theta_1} \bd \sigma_{k, \bd \theta_2}  + c_{\bd \theta_1}. \nonumber
\end{align}
Eq. \eqref{mc-qn} satisfies $\E[q^\#(\bd \theta_1 | \bd \theta_2)] = q(\bd \theta_1 | \bd \theta_2)$, where the expectation is taken with respect to $\bd \sigma_{1, \bd \theta}, \ldots, \bd \sigma_{K, \bd \theta}$.  We now define the \emph{Monte Carlo gradient} 
\begin{align}
\bd g^\#(\bd \theta) := \nabla_1 q^\#(\bd \theta | \bd \theta), \label{mc-grad}
\end{align} 
which can take the place of $\bd g^\star(\bd \theta)$ for updating $\bd \theta$ in gradient EM. 
 The estimator satisfies $\E[\bd g^\#(\bd \theta)] = \bd g^\star(\bd \theta)$.


\textbf{Constructing Samples} The question remains as to how we draw each sample $\bd \sigma_{k, \bd \theta}$.  Consider independent random vectors $\bd \xi_{k} \sim \mathcal{N}(\bd 0, \bd \Gamma_{\bd \theta})$ and $\bd \zeta_{k} \sim \mathcal{N}(\bd 0,
\bd \Psi_{\bd \theta})$, and let 
\begin{align}
\bd \delta_{k} := \bd \xi_{k} + \bd \Phi_{\bd \theta}^\top \bd \Omega^\top \bd \Omega \bd \zeta_{k}. \label{delta}
\end{align}
It follows from properties of Gaussian random vectors that $\bd \delta_{k} \sim\mathcal{N}(\bd 0, \bold A_{\bd \theta})$, where $\bold A_{\bd \theta}$ is defined in \eqref{def-a}. Then, we let 
\begin{align}
\bd \sigma_{k, \bd \theta} := \bd \Sigma_{\bd \theta} \bd \delta_{k}, \quad k = 1, \ldots, K.\label{particle}
\end{align}
As a result, $\bd \sigma_{k, \bd \theta}$ has covariance $\bold \Sigma_{\bd \theta} \bold A_{\bd \theta} \bold \Sigma_{\bd \theta} = \bold \Sigma_{\bd \theta}$.  

\subsection{Linear Systems and Iterative Solvers}
Although the large covariance matrix $\bd \Sigma_{\bd \theta}$ is no longer explicitly written in the new objective \eqref{mc-qn}, it still appears in the definitions for $\bd \mu_{\bd \theta}$ and $\bd \sigma_{k, \bd \theta}$ in \eqref{mvn} and \eqref{particle}, respectively.  
In this section, we show how to obtain $\bd \mu_{\bd \theta}, \bd \sigma_{k, \bd \theta}$ \emph{without} explicitly forming the covariance matrix. 

First, we cast $\bd \mu_{\bd \theta}$ and $\bd \sigma_{k, \bd \theta}$ as the solutions to linear systems,
\begin{align}
\bold A_{\bd \theta} \bd \mu_{\bd \theta} = \bd b_{\bd \theta}, \hspace{1.5em}
\bold A_{\bd \theta} \bd \sigma_{k, \bd \theta} = \bd \delta_{k}, \quad k = 1, \ldots, K \label{systems},
\end{align}
where $\bold A_{\bd \theta} = \bold \Sigma_{\bd \theta}^{-1}$ is defined in \eqref{def-a}. Then, we solve \eqref{systems} using an \emph{iterative linear solver} \citep{saad2003iterative}.  For a system $\bold A \bd x = \bd b$, iterative solvers refine a solution $\baidx x i$ over iterations $i = 1, \ldots, I$ until $\baidx x I \approx \bold A^{-1} \bd b$.   At iteration $i$, 
\begin{align}
\baidx x {i+1} := \baidx x {i} + \baidx p {i}, 
\end{align}
where $\baidx p i$ is the search direction. Different solvers vary in how they construct $\baidx p {i}$. Examples of popular solvers include \emph{gradient descent}, \emph{steepest descent}, and \emph{conjugate gradient} \citep{saad2003iterative}, which we review in Appendix \ref{app-solvers}.  

\subsection{Gradients from Truncated Linear Solvers}
High-dimensional latent spaces $D$ may require a large number of iterations $I$ (hence a high computational cost) to obtain exact solutions.  Thus, in practice, it is desirable to run the solver for small $I$, which leads to  approximations $(\baidx[\bd \theta] \mu I, \baidx[k, \bd \theta] \sigma I)$ of the true quantities $(\bd \mu_{\bd \theta}, \bd \sigma_{k, \bd \theta})$.  This section proposes two ways to obtain an approximate EM gradient from these partial solutions $(\baidx[\bd \theta] \mu I, \baidx[k, \bd \theta] \sigma I)$. 
We defer a theoretical analysis of the gradient error to Section~\ref{sec:thr}.

First, we substitute $(\baidx[\bd \theta] \mu I, \baidx[k, \bd \theta] \sigma I)$ for $(\bd \mu_{\bd \theta}, \bd \sigma_{k, \bd \theta})$ in \eqref{mc-qn}, i.e.
\begin{align}
q^{\langle I \rangle}(\bd \theta_1 | \bd \theta_2) := &\frac{1}{2}(\bd \mu_{\bd \theta_2}^{\langle I \rangle})^\top \bold A_{\bd \theta_1} \bd \mu_{\bd \theta_2}^{\langle I \rangle} - \bd b_{\bd \theta_1}^\top \bd \mu_{\bd \theta_2}^{\langle I \rangle}\label{q-trunc} \\
&\quad \quad + \frac{1}{2K} \sum_{k=1}^K (\bd \sigma_{k, \bd \theta_2}^{\langle I \rangle})^\top \bold A_{\bd \theta_1} \bd \sigma_{k, \bd \theta_2}^{\langle I \rangle}  + c_{\bd \theta_1}, \nonumber 
\end{align}
which satisfies $\lim_{I \to \infty} q^{\langle I \rangle}(\bd \theta_1 | \bd \theta_2) = q^\#(\bd \theta_1| \bd \theta_2)$.  

\textbf{Option 1: Output Gradient} We can take the gradient of \eqref{q-trunc} in a manner similar to \eqref{mc-grad} to obtain the \emph{output gradient} 
\begin{align}
\bd {\widehat g}^{\langle I \rangle}(\bd \theta) := \nabla_1 {q}^{\langle I \rangle} (\bd \theta | \bd \theta), \label{out-grad}
\end{align}
which satisifies $\lim_{I \to \infty} \bd {\widehat g}^{\langle I \rangle}(\bd \theta) = \bd {g}^{\#}(\bd \theta)$.
We interpret this gradient as backpropagating through only the output layer of the architecture in Fig. \ref{arch}, hence the terminology.

\textbf{Option 2: Network Gradient} Since the inputs to the output layer $(\baidx[\bd \theta] \mu I, \baidx[k, \bd \theta] \sigma I)$ are themselves functions of the parameters $\bd \theta$, a natural question arises as to the benefits of additionally propagating the gradient through these quantities (and the linear solver).  This leads to the \emph{network gradient}
\begin{align}
\bd {\widetilde g}^{\langle I \rangle}(\bd \theta) := \frac{\partial }{\partial \bd \theta} \left[{q}^{\langle I \rangle} (\bd \theta | \bd \theta) - \frac{1}{K} \sum_{k=1}^K \bd \delta_k^\top \bd \sigma_{k, \bd \theta}^{\langle I \rangle}\right], \label{net-grad}
\end{align}
which backpropagates through the whole architecture in Fig. \ref{arch}. There are two changes that \eqref{net-grad} makes to \eqref{out-grad}: (a) the use of $\frac{\partial}{\partial \bd \theta}$ instead of $\nabla_1$ means that \eqref{net-grad} differentiates with respect to both variables in \eqref{q-trunc} (not just the first argument); (b) \eqref{net-grad} has an extra term with $\bd \delta_k$, which is absent from \eqref{out-grad} but is necessary in \eqref{net-grad} to ensure $\lim_{I \to \infty} \bd {\widetilde g}^{\langle I \rangle}(\bd \theta) = \bd g^\#(\bd \theta)$  (short proof in Appendix \ref{app-net-grad-lim}; longer proof in Appendix \ref{sec:opt-proof}).  In Section \ref{opt-error}, we will show that compared to the output gradient $\bd {\widehat g}^{\langle I \rangle}$, the network gradient $\bd {\widetilde g}^{\langle I \rangle}$ exhibits a ``super-efficiency" phenomenon \citep{ablin2020super, tolooshams2022stable}, which means that it converges faster to $\bd g^\#$.

\subsection{Full Algorithm}
The probabilistic unrolling algorithm is given in Algorithm \ref{bunroll}. The \textsc{LinearSolver} step depends on the particular choice of solver; options include gradient descent, steepest descent, and conjugate gradient. In addition to circumventing matrix inversion, probabilistic unrolling provides several computational benefits over EM, which we explain below. 

\emph{Covariance-Free Computation.} The iterative solvers eliminate the need to explicitly form the $D \times D$ 
 covariance matrix $\bold \Sigma_{\bd \theta}$ (or even its inverse $\bold A_{\bd \theta}$). At each iteration $i$, a linear solver simply needs to compute matrix-vector products of the form $\bold A_{\bd \theta} \bd v$, for any $\bd v \in \R^D$, efficiently.  For an LGM, the matrix $\bold A_{\bd \theta}$ is a highly-structured function of its canonical parameters and the data mask $\bold \Omega$, as shown in \eqref{def-a}.  

\emph{Exploiting LGM Structure.} In many cases, the canonical parameters of the LGM exhibit additional structure, such as diagonal, Toeplitz, low rank, and sparse structure, to name a few examples. This can significantly reduce the computational and storage costs of each iteration of the linear solver.  For example, in applications of sparse Bayesian learning \citep{lin2022covariance}, $\bd \Phi_{\bd \theta}$ and its transpose often arise as Fourier-like operators. Efficient algorithms, in both computation and storage,  exist for applying such operators to vectors.
 For a single linear system, the time cost of the solver is $\mathcal{O}(I \tau_{\bd \theta})$, where $I$ is the number of iterations and $\tau_{\bd \theta}$ is the time needed to compute the matrix-vector multiplication $\bold A_{\bd \theta} \bd v$.  The space cost is $\mathcal{O}(D + \omega_{\bd \theta})$, where $\omega_{\bd \theta}$ is the space needed to store the canonical parameters. 
 
\emph{Amenability to Parallelization.} Iterative solvers are simple and straightforward to parallelize for solving multiple linear systems (e.g. in \eqref{systems}) through $\bold A_{\bd \theta} \bold X_{\bd \theta} = \bold B_{\bd \theta}$, where  
\begin{align}
\bold X_{\bd \theta} := 
[\bd \mu_{\bd \theta} | \bd \sigma_{1, \bd \theta} | \cdots | \bd \sigma_{K, \bd \theta}],
& &
\bold B_{\bd \theta} := 
[\bd b_{\bd \theta} | \bd \delta_{1} | \cdots | \bd \delta_{K}]. \nonumber
\vspace{-2mm}
\end{align}
For example, \citet{gardner2018gpytorch} and \citet{lin2022covariance} show how to parallelize the preconditioned conjugate gradient algorithm to solve for $\bold X_{\bd \theta}$.  They demonstrate that matrix-based parallelization is especially suitable for multi-core hardware, such as graphics processing units.   In this work, we go a step further by parallelizing the solver across data points $\{\bidx \ty n\}_{n=1}^N$ to obtain solutions $\{\boldidx[\bd \theta] X n\}_{n=1}^N$ for every $n$.  By \eqref{def-a}, the operators $\{\boldidx[\bd \theta] A n\}_{n=1}^N$ only differ in the masks $\{\bidx \Omega n\}_{n=1}^N$.  Thus, the total storage needed for performing $NK$ matrix-vector multiplications with $\{\boldidx[\bd \theta] A n\}_{n=1}^N$ is only $\mathcal{O}(NKD + \omega_{\bd \theta})$ (where $\omega_{\bd \theta}$ 
 is at most $\mathcal{O}(D^2)$) even though the matrices $\{\boldidx[\bd \theta] A n\}_{n=1}^N$ have $\mathcal{O}(N D^2)$ entries.       

 \begin{algorithm}[t]
\caption{\textsc{ProbabilisticUnrolling}} \label{bunroll}
\begin{algorithmic}[1]
\State{\textbf{Inputs:} parameters $\bd \theta$, dataset $\{\bidx \ty 1, \ldots, \bidx \ty N\}$, masks $\{\bidx \Omega 1, \ldots, \bidx \Omega N\}$, unrolling iterations $I$, samples $K$}
\MRepeat { for number of EM iterations}
    \For {$n = 1, 2, \ldots, N$}
    \State{Define $\boldidx[\bd \theta] A n$ and compute $\bidx[\bd \theta] b n$ by \eqref{def-a}.}
    \State{Draw $\bidx[1] \delta n, \ldots, \bidx[K] \delta n$ using the scheme in \eqref{delta}.}
    \State{Define $\boldidx[\bd \theta] B n \gets [\bidx[\bd \theta] b n | \bidx[1] \delta n | \ldots | \bidx[K] \delta n]$.}
    \State{$\boldidx[\bd \theta] X n \gets$ \textsc{LinearSolver}($\boldidx[\bd \theta] A n, \boldidx[\bd \theta] B n$, $I$).}
    \State{Let $[\bandaidx[\bd \theta] \mu I n | \bandaidx[1, \bd \theta] \sigma I n | \ldots | \bandaidx[K, \bd \theta] \sigma I n] \gets \boldidx[\bd \theta] X n$.}
    \If{use output gradient}
    \State{Compute gradient $\bd {\widehat g}^{\langle I \rangle, (n)}$ using \eqref{out-grad}.}
    \ElsIf{use network gradient}
    \State{Compute gradient $\bd {\widetilde g}^{\langle I \rangle, (n)}$ using \eqref{net-grad}.}
    \EndIf
    \EndFor
    \State{Update $\bd \theta \gets \bd \theta - \alpha \cdot \frac{1}{N} \sum_{n=1}^N {\bd g}^{\langle I \rangle, (n)}$.}
\EndRepeat
\end{algorithmic}
\end{algorithm}

We compare the computational complexities of gradient EM using matrix inversion and probabilistic unrolling in Table \ref{tab:complexity}.   The additional factor of $I$ in the space complexity of the network gradient comes from the need to store all $I$ intermediate states of the solver for backpropagation.

\begin{table}[t]
\caption{Comparing computational complexities of EM and PU (probabilistic unrolling). In the worst case, $\tau_{\bd \theta}$ and $\omega_{\bd \theta}$ are $\mathcal{O}(D^2)$.}
\label{tab:complexity}
\vskip 0.15in
\begin{center}
\begin{small}
\begin{tabular}{lcccc}
\toprule
 & Time & Space  \\
\midrule 
EM & $\mathcal{O}(ND^3)$ & $\mathcal{O}(ND^2)$ \\
PU (Output Gradient) & $\mathcal{O}(NKI\tau_{\bd \theta})$ & $\mathcal{O}(NKD + \omega_{\bd \theta})$ \\
PU (Network Gradient) & $\mathcal{O}(NKI \tau_{\bd \theta})$ & $\mathcal{O}(NKDI + \omega_{\bd \theta})$ \\
\bottomrule
\end{tabular}
\end{small}
\end{center}
\vskip -0.1in
\vspace{-2mm}
\end{table}




\section{Related Work}
\textbf{Efficient Learning with Linear Solvers.} Using iterative solvers to circumvent matrix inversion is a widely-known technique within numerical linear algebra \citep{saad2003iterative, halko2011finding}. Recently,  solvers such as the Lanczos algorithm \citep{lanczos1950iteration} and conjugate gradient (CG) \citep{hestenes1952methods} have become popular for accelerating gradient-based learning for Gaussian processes \citep{dong2017scalable, gardner2018gpytorch, wang2019exact, wenger2022preconditioning}.  In addition, \citet{lin2022covariance} and \citet{lin2022high} used CG to accelerate the classical EM algorithm for sparse Bayesian learning.  Many of these works consider when the number of data vectors $N = 1$, as opposed to the setting of the LGM where $N$ can be large.  They also do not consider the idea of backpropagation through the solver.  

\textbf{Backpropagating through Optimization Algorithms.} Automatic differentiation (or ``backpropagation") \cite{baydin2018automatic} has been widely used and studied in machine learning~\cite{domke2012optauto, deledalle2014stein, shaban2019truncated}. \citet{domke2012optauto} studied truncated backpropagation as a replacement for implicit differentiation \citep[e.g.][]{foo2007efficient, blondel2022efficient, bertrand2022implicit} when performing incomplete energy minimization. \citet{shaban2019truncated} studied the use of truncated backpropagation for parameter estimation using unrolled networks. Backpropagating through an unrolled parameter estimation mapping has also been applied to hyperparameter optimization~\cite{maclaurin2015gradient, franceschi2018bilevelhyp}, and constructing generative adversarial networks~\cite{metz2016unrolled}. 
 \citet{ablin2020super} theoretically studied how backpropagation can accelerate gradient estimation for bilevel (i.e. min-min) optimization problems, in the setting where the inner and outer objectives are the same, and when the inner optimization algorithm is gradient descent. Moreover, \citet{tolooshams2022stable, malezieux2021understanding} studied the acceleration phenomenon for the sparse coding problem.  This paper differs from the aforementioned prior work as follows: (a) probabilistic unrolling is designed for the specific setting of the LGM (as opposed to the general energy minimization problem of \citet{domke2012optauto}), and contains a novel Monte Carlo sampling step to avoid inversion of the covariance matrix, (b) the fact that our inner optimization originates from this sampling step
 necessitates statistical considerations and analyses absent from previous work, (c) we extend the result of \citet{ablin2020super}, showing that backpropagation can accelerate gradient estimation even in cases in which the inner and outer objectives are different, and (d) we provide gradient convergence analysis for steepest descent (an algorithm that is more sophisticated than gradient descent, requiring analysis of backpropagation through the step size).

\textbf{Unrolled Networks.} Our interpretation of unrolled solvers as a deep neural network is known as unrolled/unfolded networks in the literature. \citet{gregor2010learning} introduced this approach for solving the sparse coding problem. Prior works designed and studied deep unrolled networks~\cite{chen2018theoretical, ablin2019learning}. Moreover, unrolled networks have found advantages in various applications such as compressed sensing MRI~\citep{sun2016deep}, Poisson image denoising~\citep{tolooshams2020icml}, and pattern learning from physiological data~\citep{malezieux2021understanding}.

 \textbf{Variational EM and Variational Auto-Encoders.} Variational inference (VI) is a popular approach for approximating posterior distributions with simpler surrogates.  Using VI for the E-Step of EM leads to the \emph{variational EM (VEM)} algorithm \citep[Sec. 10.3.5]{pml2Book}, which is a potential alternative to probabilistic unrolling for accelerating EM inference. VEM is more flexible than probabilistic unrolling because it can perform inference for models outside the LGM family.  However, the most common form of VEM learns a ``mean-field" approximation to the posterior, which does not model covariance between latent variables and therefore biases the learning process away from the negative log-likelihood objective $\mathcal{L}(\bd \theta)$ \eqref{nll} \citep{lin2022covariance}; in contrast, probabilistic unrolling captures rich covariance structure using samples from the true posterior and like EM, still optimizes $\mathcal{L}(\bd \theta)$ as its central objective.    The variational auto-encoder (VAE) \citep{kingma2013auto} is one of the most widely-used instances of VEM that trains a deep neural network to perform VI.  Although VAEs are efficient tools for inference, they (a) require a separate inference network that is different from the generative model, increasing the number of parameters for training, and (b) require custom design of this network's architecture (e.g. layers, activations, etc.).  In contrast, the probabilistic unrolling architecture (Fig. \ref{arch}) is based on an interpretable linear solver that uses the same parameters as the generative model.

\section{Theoretical Analysis}
\label{sec:thr}
How well probabilistic unrolling approximates the exact EM gradient depends on the number of solver iterations, and the quality of the Monte Carlo approximation. We conduct a theoretical analysis of these two sources of error.  We begin by defining population-level quantities for each gradient,
\begin{align}
\bd h^{\star} := \frac{1}{N} \sum_{n=1}^N \bd g^{\star, (n)}  & &\bd h^{\#} := \frac{1}{N} \sum_{n=1}^N \bd g^{\#, (n)} \\
\bd {\widehat h}^{\langle I \rangle} := \frac{1}{N} \sum_{n=1}^N  \bd {\widehat g}^{\langle I \rangle, (n)} & & \bd {\widetilde h}^{\langle I \rangle} := \frac{1}{N} \sum_{n=1}^N  \bd {\widetilde g}^{\langle I \rangle, (n)} \nonumber,
\end{align}
where $\bd h^{\star}(\bd \theta) = \nabla_1\mathcal{Q}(\bd \theta| \bd \theta)$ is the exact gradient EM update of \eqref{grad-em}.  We denote the approximate gradient after $I$ iterations of probabilistic unrolling by $\bd {h}^{\langle I \rangle}$, with variants $\bd {\widehat h}^{\langle I \rangle}$ and $\bd {\widetilde h}^{\langle I \rangle}$ corresponding, respectively, to the output and network gradients defined previously. 
The quantity of interest is 
\begin{align}
\norm{\bd h^{\star} - \bd h^{\langle I \rangle}} \leq \underbrace{\norm{\bd h^{\star} - \bd h^{\#}}}_{\text{statistical error}} + \underbrace{\norm{\bd h^{\#} - \bd h^{\langle I \rangle}}}_{\text{optimization error}}, \label{overall-bound}
\vspace{-2mm}
\end{align}
which decomposes into two terms.  The first term, which we name \emph{statistical error}, comes from approximating $\bd h^{\star}$ with Monte Carlo samples. We name the second term \emph{optimization error}: this term captures the error due to performing a finite number $I$ of iterations of the linear solver. 
      
\subsection{Statistical Error}
Given $\bd \theta  \in  \Theta$, we first bound $\norm{\bd h^\star(\bd \theta) - \bd h^\#(\bd \theta)}_{\infty}$.
\begin{proposition} \label{stat-error}
Let $N$ be the number of data points, $K$ be the number of samples for each data point, and $L$ be the dimesionality of $\bd \theta$. For every $n \in \{1, \ldots, N\}$ we define
\begin{align}
\boldidx M {n, \ell} := (\bidx[\bd \theta] \Sigma n)^{1/2} \frac{\partial \boldidx[\bd \theta] A n}{\partial \theta_\ell} (\bidx[\bd \theta] \Sigma n)^{1/2},
\end{align}
where $\boldidx[\bd \theta] A n$ is defined in \eqref{def-a}, $\boldidx[\bd \theta] \Sigma n$ is defined in \eqref{mvn}, and $\frac{\partial \boldidx[\bd \theta] A n}{\partial \theta_\ell}$ is the $D \times D$ matrix of partial derivatives of the entries of $\boldidx[\bd \theta] A n$ with respect to $\theta_\ell$.  Let $\xi := \max_{\ell} \max_n \norm{\boldidx M {n, \ell}}_F$, where  $\norm{\cdot}_2$ denotes the spectral norm and $\norm{\cdot}_F$ denotes Frobenius norm. 
  
Then, there is an absolute constant $C$ such that if the number of Monte Carlo samples $K$ satisfies
\begin{align}\label{cond:K}
K \geq \frac{\log (4NL)}{C} \max_\ell \left(\frac{\max_n \norm{\boldidx M {n, \ell}}_2^2}{\sum_{n} \norm{\boldidx M {n, \ell}}_F^2}\right),
\end{align}
it follows that
\begin{align}
\Pr\left(\norm{\bd h^\star - \bd h^\#}_{\infty}  > \xi\sqrt{\frac{\log(4NL)}{CNK}}\right) \leq \frac{1}{N}.
\end{align}

\end{proposition}
We give the proof in Appendix \ref{stat-proof}.  The implication of Prop. \ref{stat-error} is that with high probability, $\bd h^\#$ is close to $\bd h^\star$.  The condition in (\ref{cond:K}) is a mild Monte Carlo sample size requirement and is satisfied for instance if $K \geq \frac{\log(4NL)}{C\max(1,\kappa N)}$, where $\kappa$ is any number such that for all $\ell,n,n'$, $\frac{\norm{\boldidx M {n, \ell}}_2^2}{\norm{\boldidx M {n', \ell}}_2^2}\geq \kappa.$


\subsection{Optimization Error}  \label{opt-error}
Next, we bound optimization error $\norm{\bd h^\#(\bd \theta) - \bd h^{\langle I \rangle}(\bd \theta)}_2$. 
 \begin{proposition} \label{prop-opt-err}
  Let $I$ denote the number of linear solver iterations.  Then, the output gradient $\baidx {\widehat h} I$ and the network gradient $\baidx {\widetilde h} I$ converge to $\bd h^\#$ with the following rates:
  \begin{align}
  \norm{\bd h^{\#} - \bd {\widehat h}^{\langle I \rangle}}_2 = \mathcal{O}(\rho^I), & &  
  \norm{\bd h^{\#} - \bd {\widetilde h}^{\langle I \rangle}}_2 = \mathcal{O}(I \rho^{2I}), \nonumber
  \end{align}
  where $\rho < 1$ is the solver convergence rate. For gradient descent (GD) and steepest descent (SD), these rates are
  \begin{align}
  \rho_\textsc{GD} := \frac{\iota - 1}{\iota}, & & \rho_\textsc{SD} := \frac{\iota - 1}{\iota + 1}, 
  \end{align}
  where $\iota$ denotes the condition number (i.e. ratio between largest and smallest eigenvalues) of the matrix $\bold A_{\bd \theta}$ \eqref{def-a}.  
 \end{proposition}

From Prop. \ref{opt-error}, we draw three conclusions: First, both the output gradient $\bd {\widehat h}^{\langle I \rangle}$ and the network gradient $\bd {\widetilde h}^{\langle I \rangle}$ converge to $\bd {h}^\#$ as $I \to \infty$. Second, $\bd {\widetilde h}^{\langle I \rangle}$ achieves asymptotically better estimation of $\bd h^{\#}$ (compared to $\bd {\widehat h}^{\langle I \rangle}$). Third, the results suggest that the error in both gradients can be decreased by the use of solvers that converge faster than gradient descent, e.g., using steepest descent (as shown in Prop. \ref{opt-error}), or conjugate gradient (CG), which has convergence rate $\rho_\text{CG} = \frac{\sqrt{\iota} - 1}{\sqrt{\iota} + 1}$~\citep{shewchuk1994introduction}.
 
The proof of Prop. \ref{prop-opt-err} is given in Appendix \ref{app-prop-opt-err}.  It relies on a connection we build between probabilistic unrolling and \emph{bilevel optimization}, i.e. minimizing functions defined as a minimum \citep{ablin2020super}.  Probabilistic unrolling \eqref{q-trunc} is an instance of bilevel optimization in which the outer level optimizes the EM objective by estimating its gradient with respect to parameters $\bd \theta$. This gradient is itself dependent on the solutions $\bd \mu_{\bd \theta}, \{\bd \sigma_{k, \bd \theta}\}_{k=1}^K$ of $K + 1$ linear systems, each equivalent to minimizing an inner quadratic function that depends on $\bd \theta$.  As part of our proof, we introduce the following two lemmas, which may be of broader interest beyond our particular setting of probabilistic unrolling for LGMs.  The first result (Lemma \ref{prop:bilevel}, proof in Appendix \ref{sec:opt-proof}) is a general statement on gradient convergence for bilevel optimization problems; it extends Prop. 2.2 of \citet{ablin2020super} to settings in which the outer and inner objectives have different forms. 
 The second result (Lemma \ref{prop:lin-solve}, proof in Appendix \ref{proof-unroll}) analyzes Jacobian convergence for iterative solvers based on gradient descent and steepest descent.         

\begin{lemma}\label{prop:bilevel}
Consider a bilevel optimization problem with outer objective $r(\bd \theta, \bd \beta)$ and inner objective  $s(\bd \theta, \bd \beta)$,
\begin{align}
\min_{\bd \theta}\ r(\bd \theta, \bd \beta^\#)\quad
\text{s.t.} \quad \bd \beta^\# := \arg\min_{\bd \beta } s(\bd \theta, \bd \beta),
\end{align}
 in which the gradients $\{\nabla_1 r(\bd \theta, \bd \beta), \nabla_2 s(\bd \theta, \bd \beta)\}$ and the second derivatives $\{\nabla_{22}^2 s(\bd \theta, \bd \beta), \nabla_{12}^2 r(\bd \theta, \bd \beta)\}$ are Lipschitz continuous in $\bd \beta$.  
 Let $\bd g^\# \coloneqq \nabla_{1} r(\bd \theta, \bd \beta^\#)$ be the desired gradient.  Let $\baidx \beta I$ denote an approximation of $\bd \beta^\#$ obtained from running an iterative (and differentiable) optimizer for $I$ steps.  We use $\baidx \beta I$ to define two approximate gradients: (1) the analytic gradient (called ``output gradient" in our work) $\bd {\widehat g}^{\langle I \rangle} := \nabla_{1} r(\bd \theta, \bd \beta^{\langle I \rangle})$ and (2) the automatic gradient (called ``network gradient" in our work) $\bd {\widetilde g}^{\langle I \rangle} := \nabla_{1} r(\bd \theta, \bd \beta^{\langle I \rangle}) + \frac{\partial \bd \beta^{\langle I \rangle}}{\partial \bd \theta} \cdot \nabla_{2} s(\bd \theta, \bd \beta^{\langle I \rangle})$.  Additionally, define the Jacobians $\bd J^{\#} \coloneqq \frac{\partial \bd \beta^\#}{\partial \bd \theta}$ and $\bd J^{\langle I \rangle} \coloneqq \frac{\partial \bd \beta^{\langle I \rangle}}{\partial \bd \theta}$, and let $\bd J^{\langle I \rangle}$ be bounded (i.e. $\norm{\bd J^{\langle I \rangle}}_2 \leq J_M$).  If the outer and inner objectives share second-order derivatives, i.e. $\nabla_{12}^2 r(\bd \theta, \bd \beta^\#) = \nabla_{12}^2 s(\bd \theta, \bd \beta^\#)$, then the analytic and automatic gradients converge at the following rates: 
\begin{align}
\norm{\bd {\widehat g}^{\langle I \rangle} - \bd g^{\#}}_2 &= \mathcal{O}(\norm{\bd \beta^{\langle I \rangle} - \bd \beta^\#}_2),\\
\norm{\bd {\widetilde g}^{\langle I \rangle} - \bd g^{\#}}_2 &= \mathcal{O}(\norm{\bd \beta^{\langle I \rangle} - \bd \beta^{\#}}_2 \cdot \norm{\bd J^{\langle I \rangle} - \bd J^{\#}}_2). \nonumber
\end{align}
\end{lemma}

\begin{lemma}\label{prop:lin-solve}
Given the bilevel optimization problem from Prop.~\ref{prop:bilevel}, let the inner objective $s(\bd \theta, \bd \beta) := \frac{1}{2} \bd \beta^{\top} \bold A_{\bd \theta} \bd \beta - \bd u_{\bd \theta}^{\top} \bd \beta$ be a strongly convex quadratic function with positive definite $\bold A_{\bd \theta}$. Given $\bd \theta$, let $\bd \beta^{\langle I \rangle} := \textsc{LinearSolver}(\bold A_{\bd \theta}, \bd u_{\bd \theta}, I)$ be the output of an $I$-step linear solver used to approximate $\bd \beta^\# := \arg \min_{\bd \beta} s(\bd \theta, \bd \beta) = \bold A_{\bd \theta}^{-1} \bd u_{\bd \theta}$. Then, for gradient descent and steepest descent as the linear solver, the Jacobian error is the following function of solver error: $\norm{\bd J^{\langle I \rangle} - \bd J^{\#}}_2 = \mathcal{O}(I \cdot \norm{\bd \beta^{\langle I \rangle} - \bd \beta^{\#}}_2)$. 
\end{lemma}

\textbf{Insights on Unrolling Depth} Taken together, Prop.~\ref{stat-error} and \ref{prop-opt-err} offer insights in choosing the number of unrolling iterations $I$. Since the overall gradient error is the sum of the optimization and statistical errors\footnote{Using the fact that the $\ell_2$-norm is an upper-bound on the $\ell_\infty$-norm, we can bound the overall error \eqref{overall-bound} in $\ell_\infty$ norm (with probability $1 - 1/N$) by adding the results of Prop. \ref{stat-error} and \ref{prop-opt-err}.}, the latter being impervious to $I$, the results suggest taking $I$ just large enough to balance the two sources of error. A rough calculation yields $I \approx C \log(NK)/\log(1/\rho)$ for output gradient and $I \approx C \log(NK)/\log(1/\rho^2)$ for network gradient, for some dimension-dependent constant $C$, where $\rho$ is the convergence rate of the solver.


\section{Experiments} \label{sec:exps}

\begin{table*}[t]
\caption{Comparing percent error in noisy AR parameter recovery and computation time for EM and probabilistic unrolling (PU).}
\label{sample-table}
\vskip 0.15in
\begin{center}
\begin{small}
\begin{tabular}{rrrrrrrrr}
\toprule
$D$ & $r(\bd \phi^\text{EM}, \bd \phi^\star)$ & $r(\bd \phi^\text{PU}, \bd \phi^\star)$ & $r(\kappa^\text{EM}, \kappa^\star)$ & $r(\kappa^\text{PU}, \kappa^\star)$  & $r(\lambda^\text{EM}, \lambda^\star)$ & $r(\lambda^\text{PU}, \lambda^\star)$ & EM Time (Best) & PU Time  \\
\midrule
$1{,}000$    &  7.5$\pm$4.7 \% & 6.8$\pm$3.1 \% & 3.6$\pm$4.4 \% & 3.1$\pm$1.5 \% & 5.0$\pm$2.9 \% & 6.0$\pm$3.7 \% & 41$\pm$0 s &  \textbf{8$\pm$0 s} \\
$3{,}000$ & 3.0$\pm$2.5 \%& 3.7$\pm$2.1 \%& 3.1$\pm$2.2 \%& 3.6$\pm$2.6 \% & 2.8$\pm$3.6 \% & 3.0$\pm$3.5 \% & 413$\pm$2 s & \textbf{10$\pm$0 s}  \\
$10{,}000$   & 1.8$\pm$1.1 \%& 2.5$\pm$2.2 \% & 1.3$\pm$0.8 \% & 1.5$\pm$0.7 \% & 1.3$\pm$0.3 \% & 1.0$\pm$0.5 \% & 1361$\pm$36 s & \textbf{29$\pm$0 s}\\
$30{,}000$   & 0.5$\pm$0.2 \%& 0.4$\pm$0.2 \%& 0.4$\pm$0.1 \%  & 0.4$\pm$0.2 \% &0.7$\pm$0.1 \% & 0.8$\pm$0.2 \% & 4139$\pm$49 s  & \textbf{87$\pm$1 s} \\
\bottomrule
\end{tabular}
\end{small}
\end{center}
\vskip -0.1in
\vspace{-1mm}
\end{table*}

We perform experiments on several LGM applications, ranging from recovering unknown parameters to solving inverse problems to predicting movie ratings. We demonstrate that probabilistic unrolling provides significant scalability over EM, without loss in model performance. In all instances of EM, we use a single gradient step for the M-Step update (i.e. gradient EM).  We implement all algorithms in PyTorch and on a single Nvidia T4 GPU with 16 GB RAM.

 The main hyperparameters for probabilistic unrolling are the number of samples $K$ and the number of solver iterations $I$.  When solving a linear system $\bold A \bd x = \bd b$, we let $I$ be just large enough so that the residual error $\norm{\bd b - \bold A \baidx x I}_2^2$ is below some small threshold (i.e. $10^{-8}$).  We set $K$ based on  our theoretical analysis \eqref{cond:K}. We keep $K$ small if either the number of data points $N$ is large or the number of parameters $L$ is small; otherwise we increase $K$.  In our experiments, we find that having $I$ and $K$ in the range [10, 30] is sufficient even when $D$ increases to (tens of) thousands of dimensions.

\subsection{Parameter Recovery for Noisy AR Models}

The noisy auto-regressive (AR) model is a time series model with applications in radar \citep{ccayir2021maximum} and  biomedical imaging \citep{luo2020improved}.  A noisy AR model of order $P$ for a time series $\bd y := \{y_d\}_{d=1}^D$ is written as
\begin{align}
&\{z_1,
\ldots,
z_P\}
 \sim \mathcal{N}(\bold 0, \bold Q_{\bd \phi}),  \quad \bold Q_{\bd \phi} \in \R^P,  \label{noisy-ar} \\
&z_d = \sum_{p=1}^P \phi_p \cdot z_{d-p} + w_d,  \quad w_d \sim \mathcal{N}(0, \kappa), \quad  P < d \leq D, \nonumber \\
&y_d = z_d + v_d, \quad \quad \quad \quad \quad \quad v_d \sim \mathcal{N}(0, \lambda), \quad 1 \leq d \leq D. \nonumber
\end{align}
The initial covariance matrix $\bold Q_{\bd \phi}$ is some function of the AR coefficients $\bd \phi := \{\phi_1, \ldots, \phi_P\}$ that ensures stationarity for the latent process (see Appendix \ref{app-ar-stat} for details).   
The model's free parameters are $\bd \theta = \{\bd \phi, \lambda, \kappa\}$.  We can write this model as an LGM \eqref{lgm},
where $\bd \nu_{\bd \theta} = \bd 0, \bd \Phi_{\bd \theta} = \bold I, \bd \eta_{\bd \theta} = \bd 0$, $\bold \Psi_{\bd \theta} = \lambda^{-1} \bold I$, and $\bold \Gamma_{\bd \theta}$ is a function of $\{\bd \phi, \kappa\}$.  

\emph{Complexity Comparison.} Using matrix inversion, exact-gradient EM will require $\mathcal{O}(D^3)$-time and $\mathcal{O}(D^2)$-space.  In comparison, probabilistic unrolling scales with the time $\tau_{\bd \theta}$ and space $\omega_{\bd \theta}$ needed for matrix-vector multiplication with the posterior inverse-covariance matrix $\bold A_{\bd \theta}$ \eqref{def-a}.  For the noisy AR model of order $P$, $\bold A_{\bd \theta}$ is a banded matrix with $2P + 1$ non-zero bands (derivation given in Appendix \ref{app-ar-stat}).  As a result, $\tau_{\bd \theta} = \mathcal{O}(DP + P^3)$ and $\omega_{\bd \theta} = \mathcal{O}(DP + P^2)$, which is much more efficient than EM.\footnote{We note that instead of using matrix inversion, we could cast \eqref{noisy-ar} as a state-space model and use a Kalman smoother to run exact-gradient EM in $\mathcal{O}(DP^3)$-time and $\mathcal{O}(DP^2)$-space (see Appendix \ref{app-kalman}).  However, unlike probabilistic unrolling, the Kalman filter is a sequential algorithm and does not parallelize across $D$.} 

\emph{Setup and Results.} We compare the accuracy and speed of exact-gradient EM and probabilistic unrolling in parameter recovery for noisy AR models of order $P = 5$.  First, we randomly sample a set of true parameters $\{\bd \phi^\star, \lambda^\star, \kappa^\star\}$, generate $N = 5$ time series according to \eqref{noisy-ar}, and randomly mask out 10\% of the observations from each time series to create $\bidx \ty 1, \ldots, \bidx \ty 5$.  Then, we perform maximum likelihood estimation using either gradient EM or probabilistic unrolling to produce parameter estimates $\{\bhat \phi, \hat{\lambda}, \hat{\kappa}\}$.  We measure accuracy using the normalized root-mean-square error (NRMSE) $r(\bd \theta, \bd \theta^\star) := \norm{\bhat \theta - \bd \theta^\star}_2 / \norm{\bd \theta^\star}_2 \times 100\%$. 
For probabilistic unrolling, we use $K = 10$ Monte Carlo samples, unroll $I = 30$ iterations of the conjugate gradient solver, and use the network gradient.  Other details can be found in Appendix \ref{ar-exp-settings}.  We report results for different values of $D$ in Table \ref{sample-table}.  Probabilistic unrolling consistently matches the performance of gradient EM, while being up to 47 times faster.  For each $D$, we report the smaller of the times between EM with matrix inversion and EM with a Kalman smoother (see Appendix \ref{app-kalman}).  Typically, inversion is faster for smaller $D$ while using the Kalman smoother is faster for larger $D$. Probabilistic unrolling is faster than both of these for all $D$.  We additionally perform comparisons between probabilistic unrolling and variational EM (as implemented through the variational auto-encoder (VAE) \citep{kingma2013auto}) in Appendix \ref{ar-vae-comp}.              

\subsection{Bayesian Compressed Sensing of Sparse Signals} \label{sec:bcs}
\begin{table}[t]
\caption{Averaged CS results (see Appendix \ref{app:breakdown} for breakdown by digit type).  Without Woodbury identity, EM time is 4725$\pm$61 s.}
\label{bcs-results}
\vskip 0.15in
\begin{center}
\begin{small}
\begin{tabular}{lcccc}
\toprule
 & $r(\bd \mu^\text{EM}, \btilde z)$ & $r(\bd \mu^\text{PU}, \btilde z)$ & EM Time & PU Time  \\
\midrule 
Avg. & 4.8$\pm$1.0 \%& 4.7$\pm$1.4 \%& 1481$\pm$19 s& \textbf{21$\pm$0 s}\\
\bottomrule
\end{tabular}
\end{small}
\end{center}
\vspace{-1em}
\vskip -0.1in
\end{table}

\begin{table*}[h]
\caption{MovieLens results.  For timing, a \emph{cycle} is defined as 2,000 gradient steps.  EM requires too much memory to run ML-25M.}
\label{fa-results}
\vskip 0.15in
\begin{center}
\begin{small}
\begin{tabular}{rrrrrrrrr}
\toprule
Dataset & $N$ (users) & $M$ (movies) & EM RMSE & PU RMSE  & EM Time/Cycle & PU Time/Cycle & PU Mem & PU Mem  \\
\midrule
ML-1M    &  6{,}000 & 4{,}000 & 0.8433 & 0.8436 & 54 min, 42 s & \textbf{5 min, 50 s} & 1.94 GB &  \textbf{0.17 GB} \\
ML-10M & 72{,}000 & 10{,}000 & 0.7809 & 0.7796 & 78 min, 36 s  & \textbf{12 min, 8 s} & 5.62 GB & \textbf{2.64 GB}  \\
ML-25M & 162{,}000 & 62{,}000 & | & 0.7700 &  | & \textbf{31 min, 11 s}&  $>$16 GB & \textbf{8.48 GB}\\
\bottomrule
\end{tabular}
\end{small}
\end{center}
\vskip -0.1in
\vspace{-2mm}
\end{table*}

With applications from radio astronomy \citep{wiaux2009compressed} to MRI \citep{lustig2008compressed}, compressed sensing (CS) is a technique for reconstructing sparse, high-dimensional signals $\bidx {\tilde z} n$ from measurements $\bidx \ty n$. \emph{Bayesian compressed sensing} \citep{ji2008bayesian, bilgic2011multi, lin2021accelerating, lin2022bayesian} is an approach to CS that employs the sparse Bayesian learning model \citep{wipf2004sparse}
\begin{align}
\bidx z n &\sim \mathcal{N}(\bd 0, \text{diag}(\bd \alpha)^{-1}),  & &n = 1, \ldots, N\label{sbl-model} \\
\bidx \ty n| \bidx z n &\sim \mathcal{N}(\bidx \Phi n \bidx z n, \beta^{-1} \bold I), &  &n=1, \ldots, N, \nonumber
\end{align}
where each $\bidx z n \in \R^D$ is an unknown signal, $\bidx {\ty} n \in \R^M$ is a measurement associated of the signal, and $\bidx \Phi n \in \R^{M \times D}$ is a so-called measurement matrix.  The free parameters $\bd \theta$ of the model are $\bd \alpha \in \R^D$ and $\beta \in \R$. When a common sparsity pattern underlies the observations $\{\bidx \ty n\}_{n=1}^N$, maximum likelihood estimation will push many of the entries $\alpha_m$ 
to adopt large values, tending to $\infty$, and, thus, encouraging sparsity of samples from the posterior $p(\bidx z n | \bidx \ty n, \bd \alpha, \beta)$ \cite{yee2017sparse}.  The mean $\bidx \mu n$ of each posterior is then used as an estimate for $\bidx {\tilde z} n$ \citep{ji2008multitask}. 

\emph{Complexity Comparison.} In several applications (e.g. MRI, astronomy), each $\bidx \Phi n = \bidx \Omega n \bd \Phi$, where $\bd \Phi \in \mathbb{C}^{D \times D}$ is the Fourier transform and $\bidx \Omega n \in \R^{M \times D}$ is a random undersampling mask.  Thus, \eqref{sbl-model} is an instance of the LGM, where $\bd \theta := \{\bd \alpha, \beta\}$.  Using gradient EM to fit $\bd \theta$ requires $\mathcal{O}(D^3)$-time and $\mathcal{O}(D^2)$-space.  On the other hand, probabilistic unrolling scales with the complexity needed to apply $\bold A_{\bd \theta}$ \eqref{def-a} to vectors; this is dominated by the Fourier transform $\bold \Phi$, which only requires $\mathcal{O}(D \log D)$-time and $\mathcal{O}(D)$-space.  

\emph{Setup and Results.} We perform CS experiments on NIST \citep{grother1995nist}, a dataset of handwritten digits.  For each digit type (i.e. 0 through 9), we sample $N = 10$ images $\bidx {\tilde z} n$ of size $128 \times 128$, which are high-dimensional signals with $D = 16{,}384$ pixels.  Each image is  naturally sparse because most pixels are zero.  For each $\bidx {\tilde z} n$, we randomly undersample its 2D Fourier transform by 15$\%$ (i.e. $M = 0.15 D$) and add noise to construct the measurement $\bidx \ty n$. Then, we fit a Bayesian compressed sensing model \eqref{sbl-model} to $\{\bidx {\tilde y} n\}_{n=1}^N$ to obtain reconstructions $\{\bidx \mu n\}_{n=1}^N$.  We measure success using the NRMSE between $\bd \mu$ and $\btilde z$, where $\bd \mu, \btilde z \in \R^{ND}$ are the concatenations of $\{\bidx \mu n\}_{n=1}^N$ and the true signals $\{\bidx {\tilde z} n\}_{n=1}^N$, respectively.  For probabilistic unrolling, we use $K = 30$ samples, $I = 25$ iterations of preconditioned conjugate gradient, and the network gradient.  More details can be found in Appendix \ref{app-bcs-details}.  
Results averaged over the 10 different digit types are given in Table \ref{bcs-results}. We find that probabilistic unrolling and gradient EM have similar error.  However, probabilistic unrolling is approximately 70 times faster than gradient EM, even after we accelerate EM using the Woodbury matrix identity (see Appendix \ref{app-woodbury}). 

\subsection{Collaborative Filtering through Factor Analysis}

The goal of recommender systems is to predict user ratings for various items.  One common approach is \emph{collaborative filtering}, in which we pool together incomplete ratings data for $M$ items across $N$ users to infer how all users would rate all items.   One of the central challenges of collaborative filtering is the inherent sparsity of the data -- for every user, we typically only observe ratings for a small fraction of items, leading to large amounts of missing data \citep{rendle2020neural, wu20211}.  In this section, we use \emph{factor analysis} models for collaborative filtering.  Factor analysis is a Bayesian analog of matrix factorization, one of the state-of-the-art methods for recommender systems \citep{koren2009matrix, lawrence2009non, rendle2019difficulty}.

Let $\bidx y n \in \R^{M}$ be the ratings for user $n$ across $M$ movies. Only part of this vector is known: $\bidx \ty n = \bidx \Omega n \bidx y n \in \R^{M_n}$, where $M_n < M$.  The factor analysis model is written as
\begin{align}
\bidx z n &\sim \mathcal{N}(\bd 0, \bold I),  \label{factor-analysis} \\
\bidx \ty n| \bidx z n &\sim \mathcal{N}(\bidx \Omega n(\bd \Phi \bidx z n + \bd \eta), \bidx \Omega  n \bd \Psi^{-1} (\bidx \Omega n)^\top),\nonumber
\end{align}
where each $\bidx z n \in \R^D$ for $D < M$ is a set of latent factors for user $n$.  The free parameters of this model are $\bd \theta := \{\bold \Phi, \bd \eta, \bd \Psi\}$, where $\bold \Phi \in \R^{M \times D}$, $\bd \eta \in \R^M$, and $\bd \Psi$ is a diagonal $M \times M$ matrix.  After estimating $\bd \theta$, we can predict any unknown rating $\idx[m] y n \not \in \bidx \ty n$ using the mean of the distribution $p(\idx[m] y n |\bidx \ty n, \bd \theta)$ (i.e. $\idx[m] {\hat y} n = \bd \phi_m^\top \bidx[\bd \theta] \mu n + \eta_m$, where $\bidx[\bd \theta] \mu n$ is defined by \eqref{mvn} and $\bd \phi_m$ is the $m$-th row of $\bold \Phi$).

\begin{figure}
\begin{center}
\centerline{\includegraphics[scale=0.4]{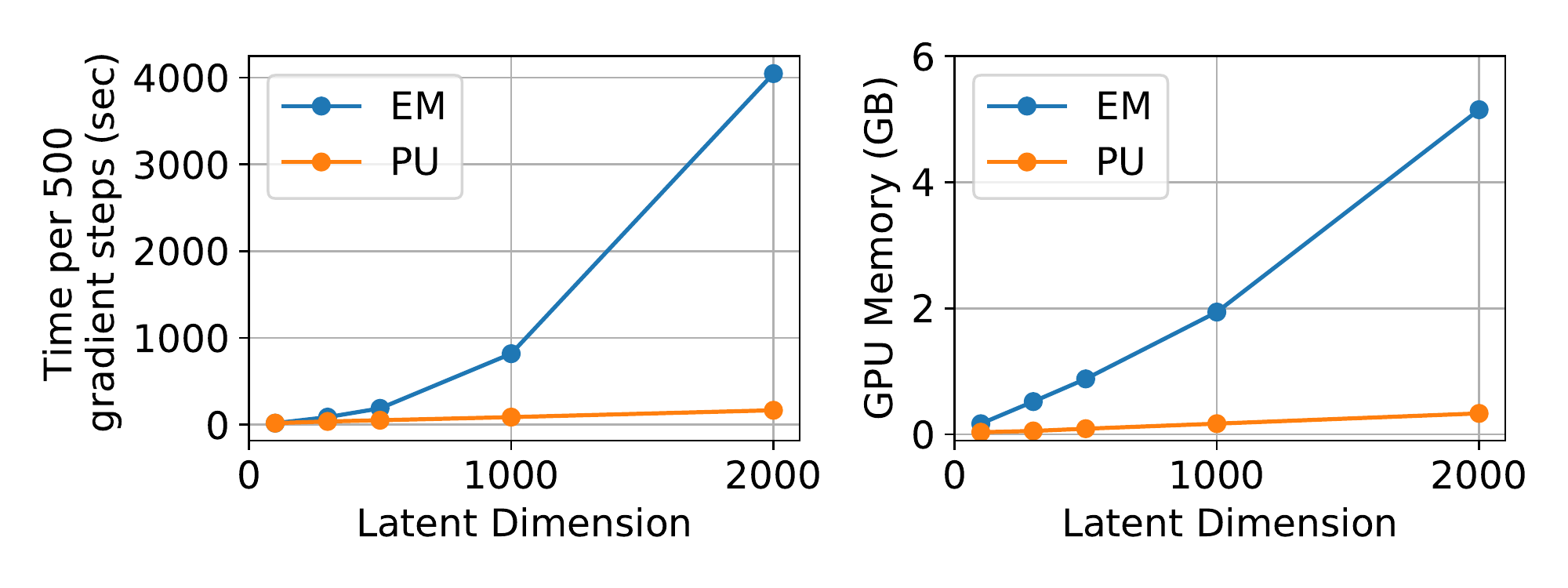}}
\vspace{-2mm}
\caption{Time and memory versus $D$ for the ML-1M dataset.}
\label{fa-scale}
\end{center}
\vspace{-3em}
\end{figure}

\emph{Setup and Results.} We perform collaborative filtering experiments on MovieLens \citep{harper2015movielens}, a group of successively larger datasets with $R = $ 1 million, 10 million, and 25 million ratings of thousands of movies, by thousands of users. For each dataset, we perform a 90\%-10\% train-test split of the ratings data \citep{sedhain2015autorec}. 
 Then, we fit a factor analysis model to the training set using mini-batch gradient descent, where the gradients are calculated using either gradient EM or probabilistic unrolling.  For probabilistic unrolling, we use $K = 10$ Monte Carlo samples, $I = 10$ unrolled iterations of conjugate gradient, and the output gradient to reduce memory consumption.  After convergence, we calculate the root-mean-square error between all ratings in the test set $\idx[m] {y} n$ and the fitted model's predictions $\idx[m] {\hat y} n$.  Further experimental details can be found in Appendix \ref{app:fa}.  The results for the three MovieLens datasets are given in Table \ref{fa-results}.  We also report processing time and GPU memory utilized by EM and probabilistic unrolling.   
 All of the results in Table \ref{fa-results} are for $D = 1{,}000$ latent factors. Figure \ref{fa-scale} shows a plot of time/memory vs. $D$ for other values of $D$.  An additional VAE baseline is provided in Appendix \ref{fa-vae-comp}.

\section{Conclusion}
We introduced \emph{probabilistic unrolling}, a computational framework for accelerating gradient-based maximum likelihood estimation for a large class of latent variable models with Gaussian prior and Gaussian likelihood.  Our method combines Monte Carlo sampling with iterative solvers and unrolled optimization, leading to a novel means of backpropagating through a sampling algorithm. Our theoretical analyses demonstrated that this can accelerate gradient estimation and, hence, maximum likelihood estimation. Our analyses provide insight into the relationship between the number of solver iterations, i.e. network depth, the number of Monte Carlo samples, and the gradient approximation error. In the future, we will consider extensions of probabilistic unrolling to other classes of probabilistic latent variable models. 

\section*{Acknowledgements}
This work was supported by a National Defense Science and Engineering Graduate Fellowship, and grants PHY-2019786, DMS-2015485, and DMS-2210664 from the National Science Foundation.
The authors also thank the anonymous reviewers, whose comments greatly improved this paper.


\bibliography{example_paper}
\bibliographystyle{icml2023}

\newpage
\appendix
\onecolumn

\section{Examples of LGMs}
\label{app:lgm-ex}
\paragraph{Factor Analysis} This model sets $\bd \nu_{\bd \theta} \gets \bd 0$ and $\bd \Gamma_{\bd \theta} \gets \bold I$ to constant values.  Given data, it learns the entries of $\bd \Phi_{\bd \theta}, \bd \eta_{\bd \theta}$ and $\bd \Psi_{\bd \theta}$ as free parameters \citep{basilevsky2009statistical}.  

\paragraph{Probabilistic PCA} Probabilistic principal components analysis is a slight variation of factor analysis in which $\bold \Psi_{\bd \theta} = \beta \bold I$ has a single free parameter $\beta$ that determines its constant diagonal values \citep{tipping1999probabilistic}.

\paragraph{Sparse Bayesian Learning}  This model assumes that $\bd \nu_{\bd \theta} \gets \bd 0$ and $\bd \eta_{\bd \theta} \gets \bd 0$.  The matrix $\bd \Phi_{\bd \theta}$ is a known and typically overcomplete (i.e. $D > M$) dictionary.  The free parameters are $\bd \theta = \{\bd \alpha, \beta\}$, where $\bd \alpha \in \R^D$ determines the prior precision $\bd \Gamma_{\bd \theta} := \text{diag}(\bd \alpha)$ and $\beta \in \R$ determines the likelihood precision $\bd \Psi_{\bd \theta} := \beta \bold I$. 
 \citep{tipping2001sparse, wipf2004sparse}.

\paragraph{State-Space Model}  This is a popular time series model that generalizes other popular variants (e.g. auto-regressive processes, moving average processes) \citep{durbin2012time}.  For a single time series $n$, it is typically written as: 
\begin{align}
\bidx[m] z n &= \bold A \bidx[m-1] z n \bd + \bidx[m] w n, && \bidx[m] w n \sim \mathcal{N}(\bd 0, \bold Q), \label{ssm}   \\
\idx[m] y n &= \bd c^\top \bidx[m] z n + \idx[m] v n, && \idx[m] v n \sim \mathcal{N}(0, \sigma^2).  \nonumber
\end{align}
At time step $m$, $\bidx[m] z n \in \R^S$ is a latent state vector and $\idx[m] y n \in \R$ is the observed data point.  We can write \eqref{ssm} in the form of \eqref{lgm} by defining $\bidx z n$ (with length $D = S \cdot M$) as the concatentation of all $\bidx[m] z n$ across $m$.  The canonical parameters can then be written as functions of the free parameters $\bd \theta := \{\bold A, \bd c, \bold Q, \sigma^2\}$.  Note that multiple time series can share these parameters in an LGM framework.       

 \paragraph{Bayesian Linear Regression}  Given a dataset of covariates and response variables $(\bd x_1, y_1), \ldots, (\bd x_M, y_M)$ where each $\bd x_m \in \R^D$, Bayesian linear regression posits the model 
 \begin{align}
 y_m = \bd x_m^\top \bd z + \varepsilon_m,
 \end{align}
 where each regression weight $z_d \sim \mathcal{N}(0, 1/\alpha)$ and each noise variable $\varepsilon_m \sim \mathcal{N}(0, 1/\beta)$ for parameters $\alpha \in \R, \beta \in \R$.  This is an LGM in which $N = 1$, $\bd \nu_{\bd \theta} \gets \bd 0$, and the rows of $\bd \Phi_{\bd \theta}$ are comprised of $\bd x_1, \ldots, \bd x_M$.  The free parameters $\bd \theta$ are $\{\alpha, \beta\}$ with $\bd \Gamma_{\bd \theta} := \alpha \bold I$ and $\bd \Psi_{\bd \theta} := \beta \bold I$ \citep{bishop2006pattern} (Section 9.3.4).

 \paragraph{Neural Linear Model}
 One modern instance of Bayesian linear regression is the \emph{neural linear model} (NLM):
  \begin{align}
 y_m = \bd f_{\bd \phi}(\bd x_m)^\top \bd z + \varepsilon_m,
 \end{align}
 where $\bd f_{\bd \phi}$ is a neural network featurizer with weights $\bd \phi$ \citep{snoek2015scalable, ober2019benchmarking}.  Thus, different from traditional Bayesian linear regression, the canonical parameter $\bold \Phi_{\bd \theta}$ is now learned through $\bd \phi$.  Therefore, the free parameteters are $\{\alpha, \beta, \bd \phi\}$.  NLMs can learn very complicated, non-linear relationships  (e.g. see Figure 1c in \citet{kristiadi2020being} and Figure 1 in \citet{ober2019benchmarking}).  The LGM also covers multi-task versions of the NLM, in which we may have multiple target vectors $\bidx y 1, \ldots, \bidx y N$ for $N > 1$ \citep{ijcai2021p334}.

\section{Derivation of $q$-Function for the LGM \eqref{qn}} \label{qstar-deriv}
Taking into account missing data with the data mask $\bidx \Omega n$, 
the model in \eqref{lgm} becomes:
\begin{align}
\bidx z n &\sim \mathcal{N}(\bd \nu_{\bd \theta}, \bd \Gamma_{\bd \theta}^{-1}),  \label{lgm-missing} \\
\bidx \ty n| \bidx z n &\sim \mathcal{N}(\bidx \Omega n(\bd \Phi_{\bd \theta} \bidx z n + \bd \eta_{\bd \theta}), \bidx \Omega  n \bd \Psi_{\bd \theta}^{-1} (\bidx \Omega n)^\top).\nonumber
\end{align}
From the definition of $q$ in \eqref{e-step}, we have (dropping the index $n$ for notational convenience)
\begin{align}
q(\bd \theta_1| \bd \theta_2) := \mathbb{E}_{p(\bd z | \bd \ty, \bd \Omega, \bd \theta_2)}[- \log p(\bd z, \bd \ty| \bd \Omega, \bd \theta_1)].
\end{align}
For any $\bd \theta \in \Theta$, the log-posterior can be written as
\begin{align}
\log p(\bd z| \bd \ty, \bd \Omega, \bd \theta) = \log p(\bd z, \bd \ty | \bd \Omega, \bd \theta) - \log p(\bd \ty| \bd \Omega, \bd \theta), 
\end{align}
where the second term is constant with respect to $\bd z$.  Expanding the first term using the probability density function for multivariate Gaussians, we have
\begin{align}
&\log p(\bd z, \bd \ty | \bd \Omega, \bd \theta) = \log p(\bd \ty| \bd z, \bd \Omega, \bd \theta) + \log p(\bd z | \bd \theta) \\
&\cong - \frac{1}{2} (\bd \ty - \bd \Omega \bd \Phi_{\bd \theta}\bd z - \bd \Omega \bd \eta_{\bd \theta})^\top (\bd \Omega \bd \Psi_{\bd \theta}^{-1} \bd \Omega^\top)^{-1} (\bd \ty - \bd \Omega \bd \Phi_{\bd \theta}\bd z - \bd \Omega \bd \eta_{\bd \theta}) + \frac{1}{2} \log \det (\bold \Omega \bold \Psi_{\bd \theta}^{-1} \bold \Omega^\top)^{-1} \nonumber \\
& \quad \quad - \frac{1}{2} (\bd z - \bd \nu_{\bd \theta})^\top \bd \Gamma_{\bd \theta} (\bd z - \bd \nu_{\bd \theta}) + \frac{1}{2} \log \det \bd \Gamma_{\bd \theta}, \nonumber
\end{align}
where $\cong$ denotes equality up to additive constants with respect to $\bd z$ and $\bd \theta$.  Note that $(\bd \Omega \bd \Psi_{\bd \theta}^{-1} \bd \Omega^\top)^{-1} = \bd \Omega \bd \Psi_{\bd \theta} \bd \Omega^\top$ because $\bd \Psi_{\bd \theta}$ is a diagonal matrix.  This leads to the simplification 
\begin{align}
&\log p(\bd z, \bd \ty | \bd \Omega, \bd \theta) \cong - \frac{1}{2} (\bd \ty - \bd \Omega \bd \Phi_{\bd \theta}\bd z - \bd \Omega \bd \eta_{\bd \theta})^\top \bd \Omega \bd \Psi_{\bd \theta} \bd \Omega^\top (\bd \ty - \bd \Omega \bd \Phi_{\bd \theta}\bd z - \bd \Omega \bd \eta_{\bd \theta}) + \frac{1}{2} \log \det (\bd \Omega \bd \Psi_{\bd \theta} \bd \Omega^\top)  \\
& \quad \quad - \frac{1}{2} (\bd z - \bd \nu_{\bd \theta})^\top \bd \Gamma_{\bd \theta} (\bd z - \bd \nu_{\bd \theta}) + \frac{1}{2} \log \det \bd \Gamma_{\bd \theta}. \nonumber
\end{align}
We can then combine terms based on whether they are quadratic, linear, or constant functions of $\bd z$ to obtain the following simplified quadratic form:
\begin{align}
\log p(\bd z, \bd \ty| \bd \Omega, \bd \theta) \cong -\frac{1}{2} \bd z^\top \bold A_{\bd \theta} \bd z +  \bd b_{\bd \theta}^\top \bd z - c_{\bd \theta}, \label{log-pdf}
\end{align}
where 
\begin{align}
\bold A_{\bd \theta} &:= \bd \Gamma_{\bd \theta} + \bd \Phi_{\bd \theta}^\top \bd \Omega^\top \bd \Omega \bd \Psi_{\bd \theta} \bd \Omega^\top \bd \Omega \bd \Phi_{\bd \theta}, \\
\bd b_{\bd \theta} &:= \bd \Gamma_{\bd \theta}  \bd \nu_{\bd \theta}  + \bd \Phi_{\bd \theta}^\top \bd \Omega^\top \bd \Omega  \bd \Psi_{\bd \theta} \bd \Omega^\top (\bd \ty - \bd \Omega \bd \eta_{\bd \theta}), \nonumber\\
c_{\bd \theta} &:= \frac{1}{2} (\bd \ty -\bd \Omega \bd \eta_{\bd \theta})^\top \bd \Omega \bd \Psi_{\bd \theta} \bd \Omega^\top (\bd \ty -\bd \Omega \bd \eta_{\bd \theta}) + \frac{1}{2} \bd \nu_{\bd \theta}^\top \bd \Gamma_{\bd \theta} \bd \nu_{\bd \theta} -  \frac{1}{2} \log \det \bold \Omega \bold \Psi_{\bd \theta} \bold \Omega^\top  - \frac{1}{2} \log \det \bd \Gamma_{\bd \theta}. \nonumber
\end{align}
By Gaussian prior-Gaussian likelihood conjugacy in \eqref{lgm}, we know that the posterior $p(\bd z| \bd \ty, \bd \Omega, \bd \theta)$ is also Gaussian with some mean $\bd \mu_{\bd \theta}$ and some covariance $\bd \Sigma_{\bd \theta}$.  From the derivations above, the log-pdf of this posterior (up to an additive constant) is given by \eqref{log-pdf}.  By matching this log-pdf to that of a $\mathcal{N}(\bd \mu_{\bd \theta}, \bd \Sigma_{\bd \theta})$, we can conclude that $\bold \Sigma_{\bd \theta} = \bold A_{\bd \theta}^{-1}$ and $\bd \mu_{\bd \theta} = \bold A_{\bd \theta}^{-1} \bd b_{\bd \theta}$.  In conclusion, we have     
\begin{align}
q(\bd \theta_1| \bd \theta_2) &= \mathbb{E}_{p(\bd z | \bd \ty, \bd \Omega, \bd \theta_2)}[- \log p(\bd z, \bd \ty| \bd \Omega, \bd \theta_1)] \cong \mathbb{E}_{\bd z \sim \mathcal{N}(\bd \mu_{\bd \theta_2}, \bd \Sigma_{\bd \theta_2})}\left[\frac{1}{2} \bd z^\top \bold A_{\bd \theta_1} \bd z -  \bd b_{\bd \theta_1}^\top \bd z + c_{\bd \theta_1}\right] \\
&= \frac{1}{2} \bd \mu_{ \bd \theta}^\top \bold A_{\bd \theta'} \bd \mu_{\bd \theta} -  \bd b_{\bd \theta'}^\top \bd \mu_{\bd \theta} + c_{\bd \theta'} + \frac{1}{2}\text{Tr}(\bold A_{\bd \theta '} \bold \Sigma_{\bd \theta}). \nonumber
\end{align}

\section{Derivation of Network Gradient Limit \eqref{net-grad}} \label{app-net-grad-lim}
We provide a short derivation, showing that the limit of the network gradient \eqref{net-grad} is the desired Monte Carlo gradient \eqref{mc-grad}.
\begin{align}
&\bd {\widetilde g}^{\langle I \rangle}(\bd \theta) := \frac{\partial }{\partial \bd \theta} \left[{q}^{\langle I \rangle} (\bd \theta | \bd \theta) - \frac{1}{K} \sum_{k=1}^K \bd \delta_k^\top \bd \sigma_{k, \bd \theta}^{\langle I \rangle}\right] \\
&=  \underbrace{\nabla_1 {q}^{\langle I \rangle} (\bd \theta | \bd \theta)}_{\bd {\widehat g}^{\langle I \rangle}(\bd \theta) \text{ by \eqref{out-grad}}} + \nabla_2 {q}^{\langle I \rangle} (\bd \theta | \bd \theta) - \frac{\partial }{\partial \bd \theta} \left[\frac{1}{K} \sum_{k=1}^K \bd \delta_k^\top \bd \sigma_{k, \bd \theta}^{\langle I \rangle}\right] \nonumber \\
&= \bd {\widehat g}^{\langle I \rangle}(\bd \theta)  + \frac{\partial \baidx[\bd \theta] \mu I}{\partial \bd \theta} \cdot \underbrace{\frac{\partial}{\partial \baidx[\bd \theta] \mu I} \left[\frac{1}{2}(\bd \mu_{\bd \theta}^{\langle I \rangle})^\top \bold A_{\bd \theta} \bd \mu_{\bd \theta}^{\langle I \rangle} - \bd b_{\bd \theta}^\top \bd \mu_{\bd \theta}^{\langle I \rangle} \right]}_{\text{converges to $\bd 0$ as $I \to \infty$}}  + \frac{1}{K} \sum_{k=1}^K \frac{\partial \baidx[k, \bd \theta] \sigma I}{\partial \bd \theta} \cdot \underbrace{\frac{\partial}{\partial \baidx[k, \bd \theta] \sigma I} \left[\frac{1}{2}(\bd \sigma_{k, \bd \theta}^{\langle I \rangle})^\top \bold A_{\bd \theta} \bd \sigma_{k, \bd \theta}^{\langle I \rangle} - \bd \delta_{k, \bd \theta}^\top \bd \sigma_{k, \bd \theta}^{\langle I \rangle} \right]}_{\text{converges to $\bd 0$ as $I \to \infty$}}\nonumber
\end{align}
The indicated derivatives converge to $\bd 0$ because $\lim_{I \to \infty} \baidx[\bd \theta] \mu I, \lim_{I \to \infty} \baidx[k, \bd \theta] \sigma I$ minimize the respective functions in brackets.  It follows that $\lim_{I \to \infty} \bd {\widetilde g}^{\langle I \rangle}(\bd \theta) = \lim_{I \to \infty} \bd {\widehat g}^{\langle I \rangle}(\bd \theta) = \bd g^\#(\bd \theta)$.


\section{Examples of Iterative Linear Solvers} \label{app-solvers}
In this appendix, we provide a few examples of iterative linear solvers for the system $\bold A \bd x = \bd b$; see \citet{saad2003iterative} for other examples.  The first example is gradient descent on the quadratic form $\frac{1}{2} \bd x^\top \bold A \bd x - \bd b^\top \bd x$, with step size parameter $\alpha \in \R$.  The next example is steepest descent, which learns the optimal step size $\aidx \alpha i$ at each iteration $i$.  Finally, we have conjugate gradient, which enforces orthogonality in the residuals $\baidx r i$ and conjugacy (with respect to $\bold A$) in the search directions $\baidx d i$.

 \begin{algorithm}[h!]
\caption{\textsc{GradientDescent}} \label{grad-desc}
\begin{algorithmic}[1]
\State{$\baidx x 0 \gets \bd 0$}
\For{$i = 1, 2, \ldots, I$}
    \State{$\baidx r i \gets \bd b - \bold A \baidx x i$}
    \State{$\baidx x i \gets \baidx x {i-1} + \alpha \cdot \baidx r i$}
\EndFor
\end{algorithmic}
\end{algorithm}

 \begin{algorithm}[h!]
\caption{\textsc{SteepestDescent}} \label{steep-desc}
\begin{algorithmic}[1]
\State{$\baidx x 0 \gets \bd 0$}
\For{$i = 1, 2, \ldots, I$}
    \State{$\baidx r i \gets \bd b - \bold A \baidx x i$}
    \State{$\aidx \alpha i \gets \dfrac{\langle \baidx r i, \baidx r i \rangle}{\langle \baidx r i, \bold A \baidx r i \rangle}$}
    \State{$\baidx x i \gets \baidx x {i-1} + \aidx \alpha i \cdot \baidx r i$}
\EndFor
\end{algorithmic}
\end{algorithm}

 \begin{algorithm}[h!]
\caption{\textsc{ConjugateGradient}} \label{conj-grad}
\begin{algorithmic}[1]
\State{$\baidx x 0 \gets \bd 0$}
\State{$\baidx r 0 \gets \bd b - \bold A \baidx x 0$}
\State{$\baidx d 0 \gets \baidx r 0$}
\For{$i = 1, 2, \ldots, I$}
    \State{$\aidx \alpha i \gets \dfrac{\langle \baidx r {i-1}, \baidx r {i-1} \rangle}{\langle \baidx d {i-1}, \bold A \baidx d {i-1} \rangle}$}
    \State{$\baidx x i \gets \baidx x {i-1} + \aidx \alpha i \cdot \baidx d {i-1}$}
    \State{$\baidx r i \gets \baidx r {i-1} + \aidx \alpha i \cdot \bold A \baidx d {i-1}$}
    \State{$\aidx \beta i \gets \dfrac{\langle \baidx r i, \baidx r i \rangle}{\langle \baidx r {i-1}, \baidx r {i-1} \rangle}$}
    \State{$\baidx d i \gets \baidx r {i} + \aidx \beta i \cdot \baidx d {i-1}$}
\EndFor
\end{algorithmic}
\end{algorithm}

\section{Proofs for Section \ref{sec:thr}}
\subsection{Proposition \ref{stat-error}} \label{stat-proof}
\begin{proof}
We will write $\bd h^\star - \bd h^\#$ as a short for $\bd h^\star(\bd \theta) - \bd h^\#(\bd \theta)$. Recall from definitions that
\begin{align}
 h^\#_\ell - h^\star_\ell = \frac{1}{N} \sum_{n=1}^N g^{\#, (n)}_\ell - g^{\star, (n)}_\ell  = \frac{1}{N} \sum_{n=1}^N [\nabla_1 q^{\#, (n)}(\bd \theta | \bd \theta) - \nabla_1 q^{(n)}(\bd \theta | \bd \theta)]_\ell
\end{align}
where $q$ is given by \eqref{qn} and $q^\#$ is given by \eqref{mc-qn}.  Observe that the only difference between these two objectives $q, q^\#$ is the trace term in $q$ versus the Monte Carlo approximation in $q^\#$. It therefore follows that for each $n$,   
\begin{align}
g^{\#, (n)}_\ell - g^{\star, (n)}_\ell  = \frob{\frac{\partial}{\partial \theta_\ell} \boldidx[\bd \theta] A n,  \frac{1}{K}\sum_{k=1}^K \bidx[k, \bd \theta] x n (\bidx[k, \bd \theta] x n)^\top -  \boldidx[\bd \theta] \Sigma n},
\end{align}
where for two matrices $\bold Q, \bold R$ we define $\frob{\bold Q, \bold R} := \text{Tr}(\bold Q \bold R^\top)$.  Recall that for all $k$, each $\bidx[k, \bd \theta] x n \sim \mathcal{N}(\bd 0, {\bidx[\bd \theta] \Sigma n})$.  Thus, we can equivalently write $\bidx[k, \bd \theta] x n = ({\bidx[\bd \theta] \Sigma n})^{1/2} \bidx[k] \epsilon n$, where each $\bidx[k] \epsilon n \sim \mathcal{N}(\bd 0, \bold I)$.  It then follows that
\begin{align}
g^{\#, (n)}_\ell - g^{\star, (n)}_\ell  &= \frob{\frac{\partial}{\partial \theta_\ell} \boldidx[\bd \theta] A n,  ({\bidx[\bd \theta] \Sigma n})^{1/2} \left(\frac{1}{K} \sum_{k=1}^K \bidx[k] \epsilon n (\bidx[k] \epsilon n)^\top -  \bold I \right)({\bidx[\bd \theta] \Sigma n})^{1/2}} \\
&= \frob{ \underbrace{({\bidx[\bd \theta] \Sigma n})^{1/2} \left(\frac{\partial}{\partial \theta_\ell} \boldidx[\bd \theta] A n\right) ({\bidx[\bd \theta] \Sigma n})^{1/2}}_{\boldidx M {n, \ell}},  \frac{1}{K} \sum_{k=1}^K \bidx[k] \epsilon n (\bidx[k] \epsilon n)^\top -  \bold I}. 
\end{align}
Our proof strategy is now similar to that of the Hanson-Wright inequality (Theorem 6.2.1 in \citet{vershynin2018high}).

Let $\idx[d, d] M {n, \ell}$ denote the $d$-th diagonal element of the $D \times D$ matrix $\boldidx M {n, \ell}$ and let $\boldidx {\widetilde M} {n, \ell}$ denote the matrix $\boldidx M {n, \ell}$ with its diagonal set to zero.  Then,  
\begin{align}
h^{\#}_\ell - h^{\star}_\ell  &= \frac{1}{NK} \sum_{n=1}^N \sum_{k=1}^K \sum_{d=1}^D \idx[d, d] M {n, \ell} \left((\idx[k, d] \epsilon n)^2 - 1\right) + \frac{1}{NK} \sum_{n=1}^N  \sum_{k=1}^K (\bidx[k] \epsilon n)^\top \boldidx {\widetilde M} {n, \ell} \bidx[k] \epsilon n.
\end{align}
We then write
\begin{align}
\Pr\left(\left| NK \cdot ( h^{\#}_\ell - h^{\star}_\ell) \right | > t\right) \leq \Pr\left(\left|\sum_{n=1}^N \sum_{k=1}^K \sum_{d=1}^D \idx[d, d] M {n, \ell} \left((\idx[k, d] \epsilon n)^2 - 1\right)\right| > \frac{t}{2}\right) \quad \quad \quad \quad \nonumber \\ 
+ \Pr\left(\left| \sum_{n=1}^N  \sum_{k=1}^K (\bidx[k] \epsilon n)^\top \boldidx {\widetilde M} {n, \ell} \bidx[k] \epsilon n \right | > \frac{t}{2}\right), \label{proof-obj}
\end{align}
and bound each of these two terms separately.  

\emph{(I) Diagonal Elements.}
We recall the definition of sub-exponential norm for a random variable $X$, 
\begin{align}
\norm{X}_{\psi_1} := \inf \left\{t > 0 : \E \left(\exp \frac{|X|}{t} \right) \leq 2 \right\}.
\end{align}
For $\idx[k, d] \epsilon n \sim \mathcal{N}(0, 1)$, we have
\begin{align}
\normv{\idx[d, d] M {n, \ell} \left((\idx[k, d] \epsilon n)^2 - 1\right)}_{\psi_1} \leq \left|\idx[d, d] M {n, \ell} \right|
 \normv{(\idx[k, d] \epsilon n)^2 - 1}_{\psi_1} \leq c_0 |\idx[d, d] M {n, \ell}|,
\end{align}
for some absolute constant $c_0$.  Then, by Bernstein's inequality, we have for all $t > 0$,
\begin{align}
\Pr\left(\left|\sum_{n=1}^N \sum_{k=1}^K \sum_{d=1}^D \idx[d, d] M {n, \ell} \left((\idx[k, d] \epsilon n)^2 - 1\right)\right| > \frac{t}{2}\right) &\leq 2 \exp \left(-c_1 \min \left(\frac{t^2}{K \sum_{n} \sum_{d=1}^D |\idx[d, d] M {n, \ell}|^2},  \frac{t}{\max_{n} \max_{d} |\idx[d, d] M {n, \ell}|}\right)\right) \nonumber \\
&\leq 2 \exp \left(-c_1 \min \left(\frac{t^2}{K \sum_{n} \norm{\boldidx M {n, \ell}}_F^2},  \frac{t}{\max_{n} \norm{\boldidx M {n, \ell}}_2}\right)\right),
\end{align}
for some absolute constant $c_1$, and where $\norm{\cdot}_F$ denotes Frobenius norm and $\norm{\cdot}_2$ denotes spectral norm.

\emph{(II) Off-diagonal Elements.}
We can bound one side of the off-diagonal term in \eqref{proof-obj} as follows: For all $\lambda \in \R$ and by Markov's inequality, we have
\begin{align}
\Pr\left(\sum_{n=1}^N  \sum_{k=1}^K (\bidx[k] \epsilon n)^\top \boldidx {\widetilde M} {n, \ell} \bidx[k] \epsilon n  > \frac{t}{2}\right) \leq \left(\exp \frac{-\lambda t}{2}\right) \prod_{n=1}^N \prod_{k=1}^K \E \left[\exp \left(\lambda \cdot( \bidx[k] \epsilon n)^\top \boldidx {\widetilde M} {n, \ell} \bidx[k] \epsilon n\right) \right]. \label{off-diag}
\end{align}
We now apply decoupling (Theorem 6.1.1 in \citet{vershynin2018high}) to have
\begin{align}
\E \left[\exp \left(\lambda \cdot( \bidx[k] \epsilon n)^\top \boldidx {\widetilde M} {n, \ell} \bidx[k] \epsilon n\right) \right] \leq \E \left[\exp \left(4\lambda \cdot( \bidx[k] \epsilon n)^\top \boldidx {\widetilde M} {n, \ell} \bidx[k] {\widetilde \epsilon} n\right) \right],
\end{align}
where $\{\bidx[k]{\widetilde \epsilon} n\}$ is an independent copy of $\{\bidx[k]{\epsilon} n\}$. Then, by a known bound on the moment generating function of Gaussian chaos (Lemma 6.2.2 in \citet{vershynin2018high}), we can find an absolute constant $c_2$, such that for $\lambda \max_n \norm{\boldidx {\widetilde M} {n, \ell}}_2 \leq c_2$,
\begin{align}
\E \left[\exp \left(4\lambda \cdot( \bidx[k] \epsilon n)^\top \boldidx {\widetilde M} {n, \ell} \bidx[k] {\widetilde \epsilon} n\right) \right] \leq \exp \left(c_2 \lambda^2 \norm{\boldidx {\widetilde M} {n, \ell}}_F^2\right).
\end{align}
Then,  
\begin{align}
\Pr\left(\sum_{n=1}^N  \sum_{k=1}^K (\bidx[k] \epsilon n)^\top \boldidx {\widetilde M} {n, \ell} \bidx[k] \epsilon n > \frac{t}{2}\right) \leq \inf_{\substack{\lambda > 0, \\ \lambda \max_n \norm{\boldidx {\widetilde M} {n, \ell}}_2 \leq c_2}} \exp \left(-\frac{\lambda t }{2} + c_2 K \lambda^2 \sum_{n=1}^N \norm{\boldidx {\widetilde M} {n, \ell}}_F^2 \right).
\end{align}
We can then optimize over $\lambda$ to get
\begin{align}
\Pr\left(\sum_{n=1}^N  \sum_{k=1}^K (\bidx[k] \epsilon n)^\top \boldidx {\widetilde M} {n, \ell} \bidx[k] \epsilon n > \frac{t}{2}\right) &\leq \exp \left(-c_2 \min \left(\frac{t^2}{K \sum_{n=1}^N \norm{\boldidx M {n, \ell}}_F^2},  \frac{t}{\max_{n} \norm{\boldidx M {n, \ell}}_2}\right)\right).
\end{align}
Bounding both sides then yields
\begin{align}
\Pr\left(\left|\sum_{n=1}^N  \sum_{k=1}^K (\bidx[k, d] \epsilon n)^\top \boldidx {\widetilde M} {n, \ell} \bidx[k, d] \epsilon n\right| > \frac{t}{2}\right) \leq 2 \exp \left(-c_2 \min \left(\frac{t^2}{ K \sum_{n=1}^N \norm{\boldidx M {n, \ell}}_F^2},  \frac{t}{\max_{n} \norm{\boldidx M {n, \ell}}_2}\right)\right).
\end{align}
Returning to our original objective in \eqref{proof-obj}, we can take $C := \min\{c_1, c_2\}$ and conclude
\begin{align}
\Pr\left(\left| NK \cdot ( h^{\#}_\ell - h^{\star}_\ell) \right | > t\right) \leq 4 \exp \left(-C \min \left(\frac{t^2}{ K \sum_{n=1}^N \norm{\boldidx M {n, \ell}}_F^2},  \frac{t}{\max_{n} \norm{\boldidx M {n, \ell}}_2}\right)\right).
\end{align}
Now, if we choose $t = \sqrt{t_0 (K / C)\sum_n \norm{\boldidx M {n, \ell}}_F^2 }$ for some $t_0 > 0$, we have
\begin{align}
\Pr\left(\left|NK \cdot (h^{\#}_\ell - h^{\star}_\ell) \right | > t\right) \leq 4 \exp \left(-C \min \left(\frac{t_0}{C},  \sqrt{\frac{\sum_n \norm{\boldidx M {n, \ell}}_F^2}{\max_{n} \norm{\boldidx M {n, \ell}}_2^2}\frac{t_0  K}{C}}\right)\right) \leq 4e^{-t_0},
\end{align}
as long as 
\begin{align}
\frac{t_0}{C} \leq \sqrt{\frac{\sum_n \norm{\boldidx M {n, \ell}}_F^2}{\max_{n} \norm{\boldidx M {n, \ell}}_2^2}\frac{t_0  K}{C}} \implies K \geq \frac{t_0}{C} \left(\frac{\max_n \norm{\boldidx M {n, \ell}}_2^2}{\sum_{n} \norm{\boldidx M {n, \ell}}_F^2}\right).
\end{align}
Since for all $\ell$, $\sum_n \norm{\boldidx M {n, \ell}}_F^2 \leq N\xi^2$,  we have
\begin{align}
\mbox{ for all }\;\ell:\;\;\; \Pr\left(\left|NK \cdot (h^{\#}_\ell - h^{\star}_\ell) \right | > \xi\sqrt{\frac{t_0 N K}{C}}\right) \leq 4 \exp(-t_0),
\end{align}
as long as 
\[K \geq \frac{t_0}{C} \max_\ell \left(\frac{\max_n \norm{\boldidx M {n, \ell}}_2^2}{\sum_{n} \norm{\boldidx M {n, \ell}}_F^2}\right).\]
By union bound and taking $t_0 = \log(4NL)$,  we can conclude that for some absolute constant $C$,
\[\Pr\left(\norm{\bd h^\star - \bd h^\#}_{\infty}  > \xi\sqrt{\frac{\log(4NL)}{CNK}}\right) \leq 4L e^{-t_0}\leq \frac{1}{N},\]
under the condition
\begin{align}\label{cond:K:proof}
K \geq \frac{\log (4NL)}{C} \max_\ell \left(\frac{\max_n \norm{\boldidx M {n, \ell}}_2^2}{\sum_{n} \norm{\boldidx M {n, \ell}}_F^2}\right).
\end{align}
Clearly, $\frac{\max_n \norm{\boldidx M {n, \ell}}_2^2}{\sum_{n} \norm{\boldidx M {n, \ell}}_F^2}\leq 1$. And if we can find $\kappa\geq 0$ such that for all $\ell,n,n'$, 
\[\frac{\norm{\boldidx M {n, \ell}}_2^2}{\norm{\boldidx M {n', \ell}}_2^2}\geq \kappa,\]
then
\[\frac{\sum_{n} \norm{\boldidx M {n, \ell}}_F^2}{\max_n \norm{\boldidx M {n, \ell}}_2^2} =\sum_{n}\frac{\norm{\boldidx M {n, \ell}}_F^2}{\norm{\boldidx M {n, \ell}}_2^2}\frac{\norm{\boldidx M {n, \ell}}_2^2}{\max_n \norm{\boldidx M {n, \ell}}_2^2} \geq \kappa N. \]
Hence the condition in (\ref{cond:K:proof}) is implied by the condition

\[K \geq \frac{\log (4NL)}{C\max(1,N\kappa)}.\]

\end{proof}

\subsection{Proposition \ref{prop-opt-err}} \label{app-prop-opt-err}
\begin{proof}
First, observe that
\begin{align}
\norm{\baidx {\widehat h} I - \bd h^\#}_2 \leq \frac{1}{N} \sum_{n=1}^N \norm{\bd {\widehat g}^{\langle I \rangle, (n)} - \bd g^{\#, (n)}}_2, & & \norm{\baidx {\widetilde h} I - \bd h^\#}_2 \leq \frac{1}{N} \sum_{n=1}^N \norm{\bd {\widetilde g}^{\langle I \rangle, (n)} - \bd g^{\#, (n)}}_2.
\end{align}
Thus, if we show that for each $n$,
\begin{align}
\norm{\bd {\widehat g}^{\langle I \rangle, (n)} - \bd g^{\#, (n)}}_2 = \mathcal{O}(\rho^I), & & \norm{\bd {\widetilde g}^{\langle I \rangle, (n)} - \bd g^{\#, (n)}}_2 = \mathcal{O}(I \cdot \rho^{2I}), \label{single-g-conv}
\end{align}
then the same convergence rates also hold for the population-level quantities $\norm{\baidx {\widehat h} I - \bd h^\#}_2$ and $\norm{\baidx {\widetilde h} I - \bd h^\#}_2$.

The rest of the proof is dedicated to proving \eqref{single-g-conv}. At a high level, we will reinterpret probabilistic unrolling as solving a bilevel optimization problem.  We will then leverage new results that we prove for general bilevel optimization (i.e. Lemmas \ref{prop:bilevel} and \ref{prop:lin-solve}) to draw conclusions about the gradients' optimization error.
 
We define the bivariate function $r: \Theta \times \R^{(K+1) \times D} \to \R$,
\begin{align}
r(\bd \theta, \{\bd \mu, \bd \sigma_1, \ldots, \bd \sigma_K\}) := &\frac{1}{2} \bd \mu^\top \bold A_{\bd \theta} \bd \mu - \bd b_{\bd \theta}^\top \bd \mu  + \frac{1}{2K} \sum_{k=1}^K \bd \sigma_{k}^\top \bold A_{\bd \theta} \bd \sigma_{k}  + c_{\bd \theta}.  \label{bilevel}
\end{align}
We now re-express the gradients $\bd g^\#, \baidx {\widehat g} I, \baidx {\widetilde g} I$ in terms of $r$ to analyze the relationships between them: For example, the Monte Carlo EM gradient \eqref{mc-grad} can be written as
\begin{align}
\bd g^\#(\bd \theta) = \nabla_{1} r(\bd \theta, \{\bd \mu_{\bd \theta}, \bd \sigma_{1, \bd \theta}, \ldots, \bd \sigma_{K, \bd \theta}\}).
\end{align}
Here, $\nabla_1 r$ denotes the gradient of $r$ with respect to its first argument $\bd \theta$ and 
\begin{align}
\bd \mu_{\bd \theta} &:= \bold A_{\bd \theta}^{-1} \bd b_{\bd \theta} =  \arg \min_{\bd \mu} \left[s_0(\bd \theta, \bd \mu) := \tfrac{1}{2} \bd \mu^\top \bold A_{\bd \theta} \bd \mu - \bd b_{\bd \theta}^\top \bd \mu \right], \label{inner-level} \\
\bd \sigma_{k, \bd \theta} &:= \bold A_{\bd \theta}^{-1} \bd u_{\bd \theta} = \arg \min_{\bd \sigma_k} \left[s_k(\bd \theta, \bd \sigma_k) := \tfrac{1}{2} \bd \sigma_k^\top \bold A_{\bd \theta} \bd \sigma_k - \bd \delta_{k}^\top \bd \sigma_k \right], \forall k, \nonumber
\end{align}
are minimizers of inner problems $\{s_k\}_{k=0}^K$  parameterized by $\bd \theta$.  Linear solvers, unrolled for $I$ iterations, approximate the minimizers with $(\baidx[\bd \theta] \mu I, \baidx[k, \bd \theta] \sigma I)$.  Using $r$ and $\{s_k\}_{k=0}^K$, we can write the output \eqref{out-grad}  and network \eqref{net-grad} gradients  as
\begin{align}
\baidx {\widehat g} I(\bd \theta) &= \nabla_1 r(\bd \theta, \{\bd \mu_{\bd \theta}^{\langle I \rangle}, \bd \sigma_{1, \bd \theta}^{\langle I \rangle}, \ldots, \bd \sigma_{K, \bd \theta}^{\langle I \rangle}\}), \\
\baidx {\widetilde g} I(\bd \theta) &= \nabla_1 r(\bd \theta, \{\bd \mu_{\bd \theta}^{\langle I \rangle}, \bd \sigma_{1, \bd \theta}^{\langle I \rangle}, \ldots, \bd \sigma_{K, \bd \theta}^{\langle I \rangle}\})  + \frac{\partial \bd \mu_{\bd \theta}^{\langle I \rangle}}{\partial \bd \theta}  \cdot \nabla_2 s_0(\bd \theta, \bd \mu_{\bd \theta}^{\langle I \rangle}) + \frac{1}{K}\sum_{k=1}^K \frac{\partial \bd \sigma_{k, \bd \theta}^{\langle I \rangle}}{\partial \bd \theta}  \cdot \nabla_2 s_k(\bd \theta, \bd \sigma_{k, \bd \theta}^{\langle I \rangle}). 
\end{align}
By Lemma \ref{prop:bilevel}, the output gradient (also called the ``analytic gradient" by \citet{ablin2020super}) and the network gradient (also called the ``automatic gradient" by \citet{ablin2020super}\footnote{Note that the analytic and automatic gradients we define here are technically more general than those defined by \citet{ablin2020super}, since \citet{ablin2020super} assume that the outer and inner optimization problems are the same (i.e. $s = r$).}) converge with rates
\begin{align}
&\norm{\bd g^{\#} - \bd {\widehat g}^{\langle I \rangle}}_2 = \mathcal{O}(\norm{\bd \beta^{\#} - \bd \beta^{\langle I \rangle}}_2), & & \norm{\bd g^{\#} - \bd {\widetilde g}^{\langle I \rangle}}_2 = \mathcal{O}(\norm{\bd J^{\langle I \rangle} - \bd J^{\#}}_2 \norm{\bd \beta^{\langle I \rangle} - \bd \beta^{\#}}_2).
\end{align}
where
\begin{align}
\bd \beta^\# := \begin{bmatrix}
\bd \mu_{\bd \theta} \\
\bd \sigma_{1, \bd \theta} \\
\vdots \\
\bd \sigma_{K, \bd \theta} \\
\end{bmatrix}, & &
\baidx \beta I := \begin{bmatrix}
\baidx[\bd \theta] \mu I \\
\baidx[1, \bd \theta] \sigma I \\
\vdots \\
\baidx[K, \bd \theta] \sigma I \\
\end{bmatrix}, & & \bd J^\# := \frac{\partial \bd \beta^\#}{\partial \bd \theta}, & & \bd J^{\langle I \rangle} := \frac{\partial \bd \beta^{\langle I \rangle}}{\partial \bd \theta}.
\end{align}
Lemma \ref{prop:lin-solve} shows that for gradient descent (GD) and steepest descent (SD) as the linear solver, 
\begin{align}
\norm{\bd \beta^{\langle I \rangle} - \bd \beta^{\#}} &= \mathcal{O}(\rho^I), \\
\norm{\bd J^{\langle I \rangle} - \bd J^{\#}} &= \mathcal{O}(I \cdot \norm{\bd \beta^{\langle I \rangle} - \bd \beta^{\#}}) = \mathcal{O}(I \cdot \rho^I),
\end{align}
where $\rho < 1$ is the solver's convergence rate.  These rates are known as \citep{saad2003iterative}
\begin{align}
  \rho_\textsc{GD} := \frac{\iota - 1}{\iota}, & & \rho_\textsc{SD} := \frac{\iota - 1}{\iota + 1}, 
  \end{align}
  where $\iota$ is the condition number of $\bold A_{\bd \theta}$.
\end{proof}


\subsection{Lemma~\ref{prop:bilevel}}\label{sec:opt-proof}

\begin{proof}
We denote the Lipschitz constant of the gradients $\nabla_1 r(\bd \theta, \bd \beta)$ and $\nabla_2 s(\bd \theta, \bd \beta)$ with respect to $\bd \beta$ by $L^r_1$ and $L^s_2$, respectively. Similarly, we let the second derivatives $\nabla_{22}^2 s(\bd \theta, \bd \beta)$ and $\nabla_{12}^2 r(\bd \theta, \bd \beta)$ be $L^s_{22}$-Lipschitz and $L^r_{12}$-Lipschitz with respect to $\bd \beta$, respectively (note that for a function $f(\bd a_1, \bd a_2)$ of two variables $\bd a_1$ and $\bd a_2$, the notation $\nabla_{ij}^2 f$ denotes the second derivative of $f$ with respect to $\bd a_i$ and $\bd a_j$ for $i, j \in \{1, 2\}$).

Recall the target gradient is
\begin{align}
\bd g^\# := \nabla_1 r(\bd \theta, \bd \beta^\#).
\end{align}
The analytic gradient is defined as
\begin{equation}
\begin{aligned}
\bd {\widehat g}^{\langle I \rangle} := \nabla_{1} r(\bd \theta, \bd \beta^{\langle I \rangle}).
\end{aligned}
\end{equation}

By Lipschitz continuity, the automatic gradient satisfies
\begin{equation}
\begin{aligned}
&\norm{\bd g^{\#} - \bd {\widehat g}^{\langle I \rangle}}_2
= \norm{ \nabla_{1} r(\bd \theta, \bd \beta^{\#}) - \nabla_{1} r(\bd \theta, \bd \beta^{\langle I \rangle})}_2
\leq L^r_{1} \norm{\bd \beta^{\#} - \bd \beta^{\langle I \rangle}}_2.
\end{aligned}
\end{equation}

Hence, the error of the analytic gradient is on the order of the approximation error of the optimizer, i.e.
\begin{equation}
\begin{aligned}
&\norm{\bd g^{\#} - \bd {\widehat g}^{\langle I \rangle}}_2 = \mathcal{O}(\norm{\bd \beta^{\#} - \bd \beta^{\langle I \rangle}}_2).
\end{aligned}
\end{equation}

Next, we bound the error of the automatic gradient,
\begin{equation}
\begin{aligned}
\bd {\widetilde g}^{\langle I \rangle} := \nabla_{1} r(\bd \theta, \bd \beta^{\langle I \rangle}) + \baidx J I \cdot \nabla_{2} s(\bd \theta, \bd \beta^{\langle I \rangle}),
\end{aligned}
\end{equation}
where $\baidx J I := \frac{\partial \bd \beta^{\langle I \rangle}}{\partial \bd \theta}$, which is assumed to be bounded $\norm{\baidx J I}_2 \leq J_M$ for some constant $J_M$.

We begin by establishing some identities that will be useful for constructing our bound.  From the inner problem, we have
\begin{equation}
\nabla_{2} s(\bd \theta, \bd \beta^{\#}) = \bd 0. \label{identity1}
\end{equation}
In addition, by the implicit function theorem, the following also holds,
\begin{equation}
\begin{aligned}
\bd J^\# := \frac{\partial \bd \beta^{\#}}{\partial \bd \theta} = - \nabla_{12}^2 s(\bd \theta, \bd \beta^{\#}) \left[\nabla_{22}^2  s(\bd \theta, \bd \beta^{\#})\right]^{-1} \implies
\bd J^{\#} \cdot \nabla_{22}^2 s(\bd \theta, \bd \beta^{\#}) + \nabla_{12}^2 s(\bd \theta, \bd \beta^{\#}) = \bd 0. \label{identity2}
\end{aligned}
\end{equation}
We have
\begin{align}
\bd g^{\#} - \bd {\widetilde g}^{\langle I \rangle}
&= \nabla_{1} r(\bd \theta, \bd \beta^{\#}) - \nabla_{1} r(\bd \theta, \bd \beta^{\langle I \rangle})
- \baidx J I \cdot \nabla_{2} s(\bd \theta, \bd \beta^{\langle I \rangle}) \\
&\quad + (\nabla_{12}^2 r(\bd \theta, \bd \beta^{\#}) - \nabla_{12}^2 r(\bd \theta, \bd \beta^{\#})) (\bd \beta^{\#} - \bd \beta^{\langle I \rangle} ) && \text{$=\bd 0$}\nonumber \\
&\quad+(\bd J^{\langle I \rangle} \cdot \nabla_{22}^2 s(\bd \theta, \bd \beta^{\#}) - \bd J^{\langle I \rangle} \cdot \nabla_{22}^2 s(\bd \theta, \bd \beta^{\#}))(\bd \beta^\# - \baidx \beta I) && \text{$=\bd 0$} \nonumber \\
&\quad+ \baidx J I \cdot \nabla_{2} s(\bd \theta, \bd \beta^{\#}) && \text{$=\bd 0$ by \eqref{identity1}}\nonumber \\
&\quad- (\bd J^{\#} \cdot \nabla_{22}^2 s(\bd \theta, \bd \beta^{\#}) + \nabla_{12}^2 s(\bd \theta, \bd \beta^{\#}))(\bd \beta^\# - \baidx \beta I) & &\text{$=\bd 0$ by \eqref{identity2}}\nonumber
\end{align}
Rearranging terms, we have
\begin{align}
\bd g^{\#} - \bd {\widetilde g}^{\langle I \rangle}
&=  (\nabla_{1} r(\bd \theta, \bd \beta^{\#}) - \nabla_{1} r(\bd \theta, \bd \beta^{\langle I \rangle})  - \nabla_{12}^2 r(\bd \theta, \bd \beta^{\#}) (\bd \beta^{\#} - \bd \beta^{\langle I \rangle} )) && \text{(Term A)} \nonumber\\
& \quad + \bd J^{\langle I \rangle} \left(\nabla_{2} s(\bd \theta, \bd \beta^{\#}) - \nabla_{2} s(\bd \theta, \bd \beta^{\langle I \rangle}) - \nabla_{22}^2 s(\bd \theta, \bd \beta^{\#}) (\bd \beta^{\#} - \bd \beta^{\langle I \rangle})\right) && \text{(Term B)} \nonumber\\
& \quad + \left(\bd J^{\langle I \rangle} \nabla_{22}^2 s(\bd \theta, \bd \beta^{\#}) + \nabla_{12}^2 r(\bd \theta, \bd \beta^{\#}) - \bd J^{\#} \nabla_{22}^2 s(\bd \theta, \bd \beta^{\#}) - \nabla_{12}^2 s(\bd \theta, \bd \beta^{\#}) \right) (\bd \beta^{\#} - \bd \beta^{\langle I \rangle}) && \text{(Term C)} \nonumber
\end{align}
We bound each term below.  From Lipschitz-continuity of the second derivative $\nabla_{12} r$, we have the quadratic bound 
\begin{equation}
\begin{aligned}
\norm{\text{(Term A)}}_2 = \norm{\nabla_1 r(\bd \theta, \bd \beta^{\#}) - \nabla_{1} r(\bd \theta, \bd \beta^{\langle I \rangle})  - \nabla_{12}^2 r(\bd \theta, \bd \beta^{\#}) (\bd \beta^{\#} - \bd \beta^{\langle I \rangle} )}_2 \leq \frac{L^r_{12}}{2} \norm{\bd \beta^{\#} - \bd \beta^{\langle I \rangle}}_2^2.
\end{aligned}
\end{equation}
Similarly, from Lipschitz-continuity of the second derivative $\nabla_{22} s$, we have another quadratic bound
\begin{equation}
\begin{aligned}
\norm{\text{(Term B)}}_2 = \norm{\bd J^{\langle I \rangle} \left(\nabla_{2} s(\bd \theta, \bd \beta^{\#}) - \nabla_{2} s(\bd \theta, \bd \beta^{\langle I \rangle}) - \nabla_{22}^2 s(\bd \theta, \bd \beta^{\#}) (\bd \beta^{\#} - \bd \beta^{\langle I \rangle})\right)}_2
\leq J_M \cdot \frac{L^s_{22}}{2} \norm{\bd \beta^{\#} - \bd \beta^{\langle I \rangle}}_2^2.
\end{aligned}
\end{equation}

Finally, 
\begin{equation}
\begin{aligned}
\norm{\text{(Term C)}}_2 &= \norm{\left( \bd J^{\langle I \rangle} \nabla_{22}^2 s(\bd \theta, \bd \beta^{\#}) + \nabla_{12}^2 r(\bd \theta, \bd \beta^{\#}) - \bd J^{\#} \nabla_{22}^2 s(\bd \theta, \bd \beta^{\#}) - \nabla_{12}^2 s(\bd \theta, \bd \beta^{\#}) \right) (\bd \beta^{\#} - \bd \beta^{\langle I \rangle})}_2\\
&= \norm{( (\bd J^{\langle I \rangle} - \bd J^{\#})  \nabla_{22}^2 s(\bd \theta, \bd \beta^{\#})
+ (\nabla_{12}^2 r(\bd \theta, \bd \beta^{\#}) - \nabla_{12}^2 s(\bd \theta, \bd \beta^{\#}))) (\bd \beta^{\#} - \bd \beta^{\langle I \rangle})}_2\\
&\leq L^s_{2} \norm{\bd J^{\langle I \rangle} - \bd J^{\#}}_2 \norm{\bd \beta^{\#} - \bd \beta^{\langle I \rangle}}_2
+ \norm{\nabla_{12}^2 r(\bd \theta, \bd \beta^{\#}) - \nabla_{12}^2 s( \bd \theta, \bd \beta^{\#})}_2 \norm{\bd \beta^{\#} - \bd \beta^{\langle I \rangle}}_2 
\end{aligned}
\end{equation}
Hence, under the sufficient condition of $s$ and $r$ sharing the same second order derivatives, i.e. $\nabla_{12}^2 r(\bd \theta, \bd \beta^{\#}) = \nabla_{12}^2 s(\bd \theta, \bd \beta^{\#})$, the automatic gradient converges as
\begin{equation}
\begin{aligned}
\norm{\bd g^{\#} - \bd {\widetilde g}^{\langle I \rangle}}_2 = \mathcal{O}(\norm{\bd J^{\langle I \rangle} - \bd J^{\#}}_2 \norm{\bd \beta^{\langle I \rangle} - \bd \beta^{\#}}_2).
\end{aligned}
\end{equation}
\end{proof}

\subsection{Lemma~\ref{prop:lin-solve}} \label{proof-unroll}

This section proves the convergence rate of the Jacobian $\baidx J I$ to $\bd J^\#$ for gradient descent and steepest descent.  See Appendix \ref{app-solvers} for a summary of these algorithms.   

\subsubsection{Solver Convergence}\label{sec:opt-code}
For completeness, we begin by proving the convergence rate of the linear solver (i.e. how fast $\baidx \beta I$ converges to $\baidx \beta \#$).
%

\paragraph{Gradient Descent} For gradient descent, we have 
\begin{equation}
\begin{aligned}
\bd \beta^{\langle I+1 \rangle} &= \bd \beta^{\langle I \rangle} - \alpha (\bold A_{\bd \theta} \bd \beta^{\langle I \rangle} - \bd u_{\bd \theta})\\
&=\bd \beta^{\langle I \rangle} - \alpha (\bold A_{\bd \theta} \bd \beta^{\langle I \rangle} - \bold A_{\bd \theta} \bd \beta^{\#}) \\
\bd \beta^{\langle I+1 \rangle} - \bd \beta^{\#} &= (\bold I - \alpha \bold A_{\bd \theta}) (\bd \beta^{\langle I \rangle} - \bd \beta^{\#})\\
\norm{\bd \beta^{\langle I+1 \rangle} - \bd \beta^{\#}} &= \norm{(\bold I - \alpha \bold A_{\bd \theta}) (\bd \beta^{\langle I \rangle} - \bd \beta^{\#})}_2\\
&\leq \norm{\bold I - \alpha \bold A_{\bd \theta}}_2 \norm{\bd \beta^{\langle I \rangle} - \bd \beta^{\#}}_2 \\
&\leq \rho_{\text{GD}} \norm{\bd \beta^{\langle I \rangle} - \bd \beta^{\#}}_2, 
\end{aligned}
\end{equation}
where $\alpha \in \R$ is the constant step size of gradient descent.  To ensure convergence, we require $\alpha \leq \frac{1}{\lambda_\text{max}(\bold A_{\bd \theta})}$, where $\lambda_\text{max}(\bold A_{\bd \theta})$ is the largest eigenvalue of matrix $\bold A_\theta$.  Assuming $\alpha = \frac{1}{\lambda_\text{max}(\bold A_{\bd \theta})}$, the spectral norm $\rho_\text{GD}$ (or largest eigenvalue) of the symmetric positive definite matrix $\bold I - \alpha \bold A_{\bd \theta}$ is 
\begin{align}
\rho_\text{GD} = 1 - \frac{\lambda_\text{min}(\bold A_{\bd \theta})}{\lambda_\text{max}(\bold A_{\bd \theta})} = \frac{\iota - 1}{\iota},
\end{align}
where $\iota := \frac{\lambda_\text{max}(\bold A_{\bd \theta})}{\lambda_\text{min}(\bold A_{\bd \theta})}$ is defined as the condition number of $\bold A_{\theta}$.
This leads to the following rate of convergence:
\begin{equation}
\norm{\bd \beta^{\langle I \rangle} - \bd \beta^{\#}}_2 = \mathcal{O} (\rho_{\text{GD}}^I).
\end{equation}

\paragraph{Steepest Descent} For steepest descent, let $\{\bd v_j\}_j$ be the set of eigenvectors of $\bold A_{\bd \theta}$, with $\norm{\bd v_j}_2 = 1$ and  corresponding eigenvalues of $\lambda_1 > \lambda_2 > \cdots > \lambda_p$. We define the error at iteration $I$ by $\bd e^{\langle I \rangle} := \bd \beta^{\langle I \rangle} - \bd \beta^{\#}$.   We express this error as a linear combination of the eigenvectors,
\begin{equation}
\bd e^{\langle I \rangle} = \sum_j \zeta^{\langle I \rangle}_j \bd v_j
\end{equation}
for some coefficients $\{\baidx[j] \zeta I\}_{j=1}^D$.
Now, we define the following residual
\begin{equation}
\bd r^{\langle I \rangle} = \bd u_{\bd \theta} - \bold A_{\bd \theta} \bd \beta^{\langle I \rangle} = - \bold A_{\bd \theta} \bd e^{\langle I \rangle} = - \sum_j \zeta^{\langle I \rangle}_j \lambda_j \bd v_j.
\end{equation}
This gives us $\norm{\bd r^{\langle I \rangle}}_2^2 = \sum_j \zeta^{\langle I \rangle 2}_j \lambda^2_j$ and $\bd r^{\langle I \rangle \top} \bold A_{\bd \theta} \bd r^{\langle I \rangle} = \sum_j \zeta^{\langle I \rangle 2}_j \lambda^3_j$. Hence, we can express the optimal step-size $\aidx \alpha I$ as
\begin{equation}
\alpha^{\langle I \rangle} = \frac{\sum_j \zeta^{\langle I \rangle 2}_j \lambda^2_j}{\sum_j \zeta^{\langle I \rangle 2}_j \lambda^3_j}.
\end{equation}
Now, we show that the updates of steepest descent are contractive,
\begin{equation}
\begin{aligned}
\norm{\bd \beta^{\langle I+1 \rangle} - \bd \beta^{\#}}_2 &= \norm{(\bold I - \alpha^{\langle I \rangle} \bold A_{\bd \theta}) (\bd \beta^{\langle I \rangle} - \bd \beta^{\#})}_2\\
&= \norm{(\bold I - \frac{\sum_j \zeta^{\langle I \rangle 2}_j \lambda^2_j}{\sum_j \zeta^{\langle I \rangle 2}_j \lambda^3_j} \bold A_{\bd \theta}) \sum_b \zeta^{\langle I \rangle}_b \bd v_b}_2\\
&= \norm{\sum_b (1 - \frac{\sum_j \zeta^{\langle I \rangle 2}_{j} \lambda^2_j}{\sum_j \zeta^{\langle I \rangle 2}_j \lambda^3_j} \lambda_b) \zeta^{\langle I \rangle}_b \bd v_b}_2 \\
&= \norm{\sum_b (1 - \frac{\sum_j \zeta^{\langle I \rangle 2}_j \lambda^2_j}{\sum_j \zeta^{\langle I \rangle 2}_j \lambda^3_j} \lambda_b) \zeta^{\langle I \rangle}_b \bd v_b}_2\\
&= \omega^{\langle I \rangle} \norm{\bd \beta^{\langle I \rangle} - \bd \beta^{\#}}_2,
\end{aligned}
\end{equation}
where
\begin{equation}
\omega^{\langle I \rangle} = \frac{\norm{\sum_b (1 - \frac{\sum_j \zeta^{\langle I \rangle 2}_j \lambda^2_j}{\sum_j \zeta^{\langle I \rangle 2}_j \lambda^3_j} \lambda_b) \zeta^{\langle I \rangle}_b \bd v_b}_2}{\norm{\sum_a \zeta^{\langle I \rangle}_a \bd v_a}_2}. 
\end{equation}
We are now left to find an upper bound on $\omega^{\langle I \rangle}$ that goes to zero as $I$ increases. We derive results for $D=2$. The condition number of $\bold A_{\bd \theta}$ is $\iota \coloneqq \frac{\lambda_1}{\lambda_2} > 1$. We denote $\tau^{\langle I \rangle} \coloneq = \frac{\zeta^{\langle I \rangle}_2}{\zeta^{\langle I \rangle}_1}$. We have
\begin{equation}
\begin{aligned}
\omega^{\langle I \rangle} &= \frac{\norm{\sum_b (1 - \frac{\sum_j \zeta^{\langle I \rangle 2}_j \lambda^2_j}{\sum_j \zeta^{\langle I \rangle 2}_j \lambda^3_j} \lambda_b) \zeta^{\langle I \rangle}_b \bd v_b}_2 }{\norm{ \sum_a \zeta^{\langle I \rangle}_a \bd v_a}_2}\\
&= \frac{\norm{\sum_b (1 - \frac{\zeta^{\langle I \rangle 2}_1 \lambda^2_1 + \zeta^{\langle I \rangle 2}_2 \lambda^2_2}{\zeta^{\langle I \rangle 2}_1 \lambda^3_1 + \zeta^{\langle I \rangle 2}_2 \lambda^3_2} \lambda_b) \zeta^{\langle I \rangle}_b \bd v_b}_2 }{\norm{  \zeta^{\langle I \rangle}_1 \bd v_1 + \zeta^{\langle I \rangle}_2 \bd v_2}_2}\\
&= \frac{\norm{\sum_b (1 - \frac{\iota^2 + \tau^{\langle I \rangle 2}}{\lambda_2 (\iota^3 + \tau^{\langle I \rangle 2})} \lambda_b) \zeta^{\langle I \rangle}_b \bd v_b}_2}{\norm{\zeta^{\langle I \rangle}_1 \bd v_1 + \zeta^{\langle I \rangle}_2 \bd v_2}_2}\\
&= \frac{\norm{(1 - \frac{\iota^2 + \tau^{\langle I \rangle 2} }{\lambda_2 (\iota^3 + \tau^{\langle I \rangle 2})} \lambda_1) \zeta^{\langle I \rangle}_1 \bd v_1 + (1 - \frac{\iota^2 + \tau^{\langle I \rangle 2} }{\lambda_2 (\iota^3 + \tau^{\langle I \rangle 2})} \lambda_2) \zeta^{\langle I \rangle}_2 \bd v_2}_2}{\norm{\zeta^{\langle I \rangle}_1 \bd v_1 + \zeta^{\langle I \rangle}_2 \bd v_2}_2}\\
&= \frac{\norm{(\zeta^{\langle I \rangle}_1 \bd v_1 + \zeta^{\langle I \rangle}_2 \bd v_2) - (\iota \frac{\iota^2 + \tau^{\langle I \rangle 2} }{(\iota^3 + \tau^{\langle I \rangle 2})}) \zeta^{\langle I \rangle}_1 \bd v_1 - (\frac{\iota^2 + \tau^{\langle I \rangle 2}}{(\iota^3 + \tau^{\langle I \rangle 2})}) \zeta^{\langle I \rangle}_2 \bd v_2}_2}{\norm{\zeta^{\langle I \rangle}_1 \bd v_1 + \zeta_{i2} \bd v_2}_2}\\
&= \frac{\norm{(\zeta^{\langle I \rangle}_1 \bd v_1 + \zeta^{\langle I \rangle}_2 \bd v_2) - \frac{\iota^2 + \tau^{\langle I \rangle 2} }{(\iota^3 + \tau^{\langle I \rangle 2})} (\iota \zeta^{\langle I \rangle}_1 \bd v_1 + \zeta^{\langle I \rangle}_2 \bd v_2)}_2}{\norm{\zeta^{\langle I \rangle}_1 \bd v_1 + \zeta^{\langle I \rangle}_2 \bd v_2}_2}\\
&= \frac{\norm{(\bd v_1 + \tau^{\langle I \rangle} \bd v_2) - \frac{\iota^2 + \tau^{\langle I \rangle 2} }{(\iota^3 + \tau^{\langle I \rangle 2})} (\iota \bd v_1 + \tau^{\langle I \rangle} \bd v_2)}_2}{\norm{\bd v_1 + \tau^{\langle I \rangle} \bd v_2}_2}.\\
\end{aligned}
\end{equation}
The worst convergence (i.e. an upper bound) is achieved when $\iota = \tau^{\langle I \rangle}$. Hence, we write the upper bound on $\omega^{\langle I \rangle}$ as follows:
\begin{equation}
\begin{aligned}
\omega^{\langle I \rangle} &= \frac{\norm{(\bd v_1 + \tau^{\langle I \rangle} \bd v_2) - \frac{\iota^2 + \tau^{\langle I \rangle 2} }{(\iota^3 + \tau^{\langle I \rangle 2})} (\iota \bd v_1 + \tau^{\langle I \rangle} \bd v_2)}_2}{\norm{\bd v_1 + \tau^{\langle I \rangle} \bd v_2}_2}\\
&\leq \frac{\norm{(\bd v_1 + \iota \bd v_2) - \frac{\iota^2 + \iota^2 }{(\iota^3 + \iota^2)} (\iota \bd v_1 + \iota \bd v_2)}_2}{\norm{\bd v_1 + \iota \bd v_2}_2}\\
&= \frac{\norm{(\bd v_1 + \iota \bd v_2) - \frac{2 \iota^2}{\iota^2 (1 + \iota)} (\iota \bd v_1 + \iota \bd v_2)}_2}{\norm{\bd v_1 + \iota \bd v_2}_2}\\
&= \frac{\norm{(\bd v_1 + \iota \bd v_2) - \frac{2 \iota}{1 + \iota} (\bd v_1 + \bd v_2)}_2}{\norm{\bd v_1 + \iota \bd v_2}_2}\\
&= \frac{\norm{\frac{1 - \iota}{\iota + 1} \bd v_1 + \frac{\iota (\iota - 1)}{\iota + 1} \bd v_2}_2}{\norm{\bd v_1 + \iota \bd v_2}_2}\\
&= \frac{ \frac{\iota - 1}{\iota + 1} \norm{\bd v_1 + \iota  \bd v_2}_2}{\norm{\bd v_1 + \iota \bd v_2}_2}\\
&= \frac{\iota - 1}{\iota + 1}.
\end{aligned}
\end{equation}
Hence,
\begin{equation}
\norm{\bd \beta^{\langle I + 1 \rangle} - \bd \beta^{\#}}_2 \leq \left(\frac{\iota - 1}{\iota + 1}\right) \norm{\bd \beta^{\langle I \rangle} - \bd \beta^{\#}}_2.  
\end{equation}
We denote $\rho_{\text{SD}} \coloneqq \frac{\iota - 1}{\iota + 1}$ (which is a faster rate than $\rho_\text{GD}$), and write
\begin{equation}
\norm{\bd \beta^{\langle I \rangle} - \bd \beta^{\#}}_2 = \mathcal{O}(\rho_{\text{SD}}^I).
\end{equation}
This bound also holds for $D>2$; the proof steps are similar (e.g. see Section 9.2 of~\citet{shewchuk1994introduction}).
\subsubsection{Jacobian Convergence}\label{sec:opt-jac}
We now analyze the convergence rate of the Jacobian. 
 Convergence studies in this context date back to the seminal work of \citet{gilbert1992automatic}. We drop the subscript $\bd \theta$ when referring to $(j,k)$-entry of $\bold A_{\bd \theta}$ for ease of notation. Given matrix $\bold A$, we have
\begin{equation}
\partial \lambda_j = \bd v_j^{\top} \partial \bold A_{\bd \theta} \bd v_j
\end{equation}
\begin{equation}
\partial \bd v_j = (\lambda_j \bold I - \bold A_{\bd \theta})^{\dag} \partial \bold A_{\bd \theta} \bd v_j
\end{equation}
\begin{equation}
\frac{\partial \bold A_{\bd \theta}}{\partial \bold A_{jk}} = \bold H^{jk}, 
\end{equation}
where $\dag$ denotes pseudo inverse and $\bold H$ is a zero matrix except at the $(j, k)$-entry, which is $1$. The above holds when the eigenvalues and vectors are distinct. We denote the Jacobian error by $\bold B^{\langle I \rangle} := \bd J^{\langle I \rangle} - \bd J^{\#} = \frac{\partial \bd \beta^{\langle I \rangle}}{\partial \bd \theta} - \frac{\partial \bd \beta^{\#}}{\partial \bd \theta}$. Given the expression $\bd \beta^{\langle I \rangle} - \bd \beta^{\#} = \sum_b \zeta^{\langle I \rangle}_b \bd v_b$, we first focus on writing the Jacobian error with respect to $\bold A_{jk}$, denoted by $\bold B_{\bold A_{jk}}^{\langle I \rangle} = \frac{\partial \bd e^{\langle I \rangle}}{\partial \bold A_{jk}}$. Then, we use this Jacobian error to find the bound on the Jacobian with respect to $\bd \theta$ as $\bold B^{\langle I \rangle}  = \sum_{jk} \frac{\partial \bold A_{jk}}{\partial \bd \theta} \frac{\partial \bd e^{\langle I \rangle}}{\partial \bold A_{jk}}$. 

\paragraph{Gradient Descent}
For gradient descent, we start with the recursion below,
\begin{equation}
\begin{aligned}
\bd \beta^{\langle I+1 \rangle} - \bd \beta^{\#} &= (\bold I - \alpha \bold A_{\bd \theta}) (\bd \beta^{\langle I \rangle} - \bd \beta^{\#}).
\end{aligned}
\end{equation}
Then, we take the derivative with respect to $\bold A_{jk}$,
\begin{equation}
\begin{aligned}
\bold B^{\langle I+1 \rangle}_{\bold A_{jk}} &= (\bold I - \alpha \bold A_{\bd \theta}) \bold B^{\langle I \rangle}_{\bold A_{jk}} + \frac{\partial (\bold I - \alpha \bold A_{\bd \theta})}{\partial \bold A_{jk}} \bd e^{\langle I \rangle}\\
&= (\bold I - \alpha \bold A_{\bd \theta}) \bold B^{\langle I \rangle}_{\bold A_{jk}} - \alpha \bold H^{jk} \bd e^{\langle I \rangle}.\\
\end{aligned}
\end{equation}
We use the eigendecomposition of $\bold A_{\bd \theta} = \bold V \bold \Sigma \bold V^{\top}$ (i.e. $\bold V$ is the eigenvectors of $\bold A_{\bd \theta}$, and $\bold \Sigma$ is a diagonal matrix of $\bold A$ eigenvalues.This leads to the bound
\begin{equation}
\begin{aligned}
\norm{\bold B^{\langle I+1 \rangle}_{\bold A_{jk}}}_2 &\leq \norm{\bold I - \alpha \bold A_{\bd \theta}}_2 \norm{\bold B^{\langle I \rangle}_{\bold A_{jk}}}_2 + \norm{\alpha \bold H^{jk} \bd e^{\langle I \rangle}}_2\\
&\leq \rho_{\text{GD}} \norm{\bold B^{\langle I \rangle}_{\bold A_{jk}}}_2 + \norm{\alpha \bold H^{jk}}_2 \norm{\bd e^{\langle I \rangle}}_2.\\
\end{aligned}
\end{equation}
Unrolling the recursion gives us
\begin{equation}
\begin{aligned}
\norm{\bold B^{\langle I \rangle}_{\bold A_{jk}}}_2 = \mathcal{O}(I \rho_{\text{GD}}^I).
\end{aligned}
\end{equation}
Given $\bold B^{\langle I \rangle}  = \sum_{jk} \frac{\partial \bold A_{jk}}{\partial \bd \theta} \bold B^{\langle I \rangle}_{\bold A_{jk}}$, for gradient descent, we have
\begin{equation}
\begin{aligned}
\norm{\bold B^{\langle I \rangle}}_2 = \mathcal{O}(I \rho_{\text{GD}}^I).
\end{aligned}
\end{equation}
\paragraph{Steepest Descent} For steepest descent, we analyze Jacobian convergence under two different scenarios: (a) the gradient with respect to $\bd \theta$ is \emph{not} propagated through the adaptive step size $\aidx \alpha I$, and (b) the gradient is propagated through $\aidx \alpha I$ (i.e. a more sophisticated scenario). In both of these scenarios, we show that the Jacobian converges at the same asymptotic rate. We express the Jacobian error as
\begin{equation}
\begin{aligned}
\bold B_{\bold A_{jk}}^{\langle I \rangle}  &= \sum_b \zeta^{\langle I \rangle}_b \frac{\partial \bd v_b}{\partial \bold A_{jk}}\\
&= \sum_b \zeta^{\langle I \rangle}_b (\lambda_b \bd I - \bold A_{\bd \theta})^{\dag} \bold H^{jk} \bd v_b\\
&= \sum_b \zeta^{\langle I \rangle}_b (\lambda_b \bd I - \bold V \bold \Sigma \bold V^{\top})^{\dag} \bold H^{jk} \bd v_b\\
&= \sum_b \zeta^{\langle I \rangle}_b (\bold V \bold \Lambda_b \bold V^{\top}) \bold H^{jk} \bd v_b\\
&= \sum_b \sum_d \zeta^{\langle I \rangle} _b (q_{bd} \bd v_{jd} \bd v_{bk}) \bd v_d,
\end{aligned}
\end{equation}
where $\bold \Lambda_b$ is a diagonal matrix with $q_{bd} := \frac{1}{\lambda_b - \lambda_d}$ is its $d$-th diagonal entry, and $0$ on $b$-th diagonal entry. We substitute this expression into the Jacobian recursion,
\begin{equation}
\begin{aligned}
\bold B^{\langle I+1 \rangle}_{\partial \bold A_{jk}} &= (\bold I - \alpha^{\langle I \rangle} \bold A_{\bd \theta}) \bold B^{\langle I \rangle}_{\bold A_{jk}} + \frac{\partial (\bold I - \alpha^{\langle I \rangle} \bold A_{\bd \theta})}{\partial \bold A_{jk}} \bd e^{\langle I \rangle}.
\end{aligned}
\end{equation}

\emph{Scenario (a).} 
\begin{equation}
\begin{aligned}
\bold B^{\langle I+1 \rangle}_{\bold A_{jk}} &= (\bold I - \alpha^{\langle I \rangle} \bold A_{\bd \theta}) \bold B^{\langle I \rangle}_{\bold A_{jk}} - \alpha^{\langle I \rangle} \bold H^{jk} \bd e^{\langle I \rangle}\\
&= (\bold I - \alpha^{\langle I \rangle} \bold A_{\bd \theta}) \sum_b \sum_d \zeta^{\langle I \rangle}_b (q_{bd} \bd v_{jd} \bd v_{bk}) \bd v_d - \alpha^{\langle I \rangle} \bold H^{jk} \bd e^{\langle I \rangle}\\
&= \sum_b \sum_d \zeta^{\langle I \rangle}_b (q_{bd} \bd v_{jd} \bd v_{bk}) (\bold I - \alpha^{\langle I \rangle} \bold A_{\bd \theta}) \bd v_d - \alpha^{\langle I \rangle} \bold H^{jk} \bd e^{\langle I \rangle}\\
&= \sum_b \sum_d \zeta^{\langle I \rangle}_b (q_{bd} \bd v_{jd} \bd v_{bk}) (1 - \alpha^{\langle I \rangle} \lambda_d) \bd v_d - \alpha^{\langle I \rangle} \bold H^{jk} \bd e^{\langle I \rangle}\\
&= \sum_b \sum_d \zeta^{\langle I \rangle}_b (q_{bd} \bd v_{jd} \bd v_{bk}) (1 - \frac{\sum_o \zeta^{\langle I \rangle 2}_o \lambda^2_o}{\sum_o \zeta^{\langle I \rangle 2}_o \lambda^3_o} \lambda_d) \bd v_d - \alpha^{\langle I \rangle} \bold H^{jk} \bd e^{\langle I \rangle}.\\
\end{aligned}
\end{equation}
Similar to the analysis for the solver error, we consider the case when $D=2$.  We have
\begin{equation}
\begin{aligned}
\bold B^{\langle I+1 \rangle}_{\bold A_{jk}} &= \sum_b \sum_d \zeta^{\langle I \rangle}_b (q_{bd} \bd v_{jd} \bd v_{bk}) \bd v_d (1 - \frac{\sum_o \zeta^{\langle I \rangle 2}_o \lambda^2_o}{\sum_o \zeta^{\langle I \rangle 2}_o \lambda^3_o} \lambda_d) - \alpha^{\langle I \rangle} \bold H^{jk} \bd e^{\langle I \rangle}\\
&= \sum_b \sum_d \zeta^{\langle I \rangle}_b (q_{bd} \bd v_{jd} \bd v_{bk}) \bd v_d (1 - \frac{\sum_o \zeta^{\langle I \rangle 2}_o \lambda^2_o}{\sum_o \zeta^{\langle I \rangle 2}_o \lambda^3_o} \lambda_d) - \alpha^{\langle I \rangle} \bold H^{jk} \bd e^{\langle I \rangle}\\
&= \sum_b \left(\zeta^{\langle I \rangle}_b (q_{b1} \bd v_{j1} \bd v_{bk}) \bd v_1 (1 - \frac{\zeta^{\langle I \rangle 2}_1 \lambda^2_1 + \zeta^{\langle I \rangle 2}_2 \lambda^2_2}{\zeta^{\langle I \rangle 2}_1 \lambda^3_1 + \zeta^{\langle I \rangle 2}_2 \lambda^3_2} \lambda_1) + \zeta^{\langle I \rangle}_b (q_{b2} \bd v_{j2} \bd v_{bk}) \bd v_2 (1 - \frac{\zeta^{\langle I \rangle 2}_1 \lambda^2_1 + \zeta^{\langle I \rangle 2}_2 \lambda^2_2}{\zeta^{\langle I \rangle 2}_1 \lambda^3_1 + \zeta^{\langle I \rangle 2}_2 \lambda^3_2} \lambda_2) \right) - \alpha^{\langle I \rangle} \bold H^{jk} \bd e^{\langle I \rangle}\\
&= \sum_b \left(\zeta^{\langle I \rangle}_b (q_{b1} \bd v_{j1} \bd v_{bk}) \bd v_1 (1 - \frac{\iota^2 + \tau^{\langle I \rangle 2}}{\lambda_2 (\iota^3 + \tau^{\langle I \rangle 2})} \lambda_1) + \zeta^{\langle I \rangle}_b (q_{b2} \bd v_{j2} \bd v_{bk}) \bd v_2 (1 - \frac{\iota^2 + \tau^{\langle I \rangle 2}}{\lambda_2 (\iota^3 + \tau^{\langle I \rangle 2})} \lambda_2) \right) - \alpha^{\langle I \rangle} \bold H^{jk} \bd e^{\langle I \rangle}\\
&= \sum_b \left(\zeta^{\langle I \rangle}_b (q_{b1} \bd v_{j1} \bd v_{bk}) \bd v_1 (1 - \frac{\iota^2 + \tau^{\langle I \rangle 2}}{\iota^3 + \tau^{\langle I \rangle 2}} \iota) + \zeta^{\langle I \rangle}_b (q_{b2} \bd v_{j2} \bd v_{bk}) \bd v_2 (1 - \frac{\iota^2 + \tau^{\langle I \rangle 2}}{\iota^3 + \tau^{\langle I \rangle 2}}) \right) - \alpha^{\langle I \rangle} \bold H^{jk} \bd e^{\langle I \rangle}.\\
\end{aligned}
\end{equation}
Now, for the upper bound, we set $\iota = \tau^{\langle I \rangle}$
\begin{equation}
\begin{aligned}
\norm{\bold B^{\langle I+1 \rangle}_{\bold A_{jk}}}_2 &=  \norm{\sum_b \left(\zeta^{\langle I \rangle}_b (q_{b1} \bd v_{j1} \bd v_{bk}) \bd v_1 (1 - \frac{\iota^2 + \tau^{\langle I \rangle 2}}{\iota^3 + \tau^{\langle I \rangle 2}} \iota) + \zeta^{\langle I \rangle}_b (q_{b2} \bd v_{j2} \bd v_{bk}) \bd v_2 (1 - \frac{\iota^2 + \tau^{\langle I \rangle 2}}{\iota^3 + \tau^{\langle I \rangle 2}}) \right) - \alpha^{\langle I \rangle} \bold H^{jk} \bd e^{\langle I \rangle}}_2\\
&\leq \norm{\sum_b \left(\zeta^{\langle I \rangle}_b (q_{b1} \bd v_{j1} \bd v_{bk}) \bd v_1 (1 - \frac{\iota^2 + \tau^{\langle I \rangle 2}}{\iota^3 + \tau^{\langle I \rangle 2}} \iota) + \zeta^{\langle I \rangle}_b (q_{b2} \bd v_{j2} \bd v_{bk}) \bd v_2 (1 - \frac{\iota^2 + \tau^{\langle I \rangle 2}}{\iota^3 + \tau_i^2}) \right)}_2 + \norm{\alpha^{\langle I \rangle} \bold H^{jk} \bd e^{\langle I \rangle}}_2\\
&\leq \norm{\sum_b \left(\zeta^{\langle I \rangle}_b (q_{b1} \bd v_{j1} \bd v_{bk}) \bd v_1 (1 - \frac{2}{1 + \iota} \iota) + \zeta^{\langle I \rangle}_b (q_{b2} \bd v_{j2} \bd v_{bk}) \bd v_2 (1 - \frac{2}{1 + \iota}) \right)}_2 + \norm{ \alpha^{\langle I \rangle} \bold H^{jk} \bd e^{\langle I \rangle}}_2\\
&\leq \norm{\sum_b \left(\zeta^{\langle I \rangle}_b (q_{b1} \bd v_{j1} \bd v_{bk}) \bd v_1 (\frac{1 - \iota}{\iota + 1}) + \zeta^{\langle I \rangle}_b (q_{b2} \bd v_{j2} \bd v_{bk}) \bd v_2 (\frac{\iota - 1}{\iota + 1}) \right)}_2 + \norm{\alpha_i \bold H^{jk} \bd e^{\langle I \rangle}}_2\\
&= \norm{\frac{\iota - 1}{\iota + 1} \sum_b \left(\zeta^{\langle I \rangle}_b (q_{b1} \bd v_{j1} \bd v_{bk}) \bd v_1 + \zeta^{\langle I \rangle}_b (q_{b2} \bd v_{j2} \bd v_{bk}) \bd v_2 \right)}_2 + \norm{\alpha^{\langle I \rangle} \bold H^{jk} \bd e^{\langle I \rangle}}_2\\
&= \frac{\iota - 1}{\iota + 1} \norm{\bold B^{\langle I \rangle}_{\bold A_{jk}}}_2 + \norm{\alpha^{\langle I \rangle} \bold H^{jk} \bd e^{\langle I \rangle}}_2.\\
\end{aligned}
\end{equation}
Unrolling the recursion, we have
\begin{equation}
\norm{\bold B^{\langle I \rangle}_{\bold A_{jk}}} = \mathcal{O}(I \rho_{\text{SD}}^I).
\end{equation}
Hence, 
\begin{equation}
\norm{\bold B^{\langle I \rangle}}_2 = \mathcal{O}(I \rho_{\text{SD}}^I).
\end{equation}

\emph{Scenario (b).} In scenario (a), we have $\frac{(\bold I - \alpha^{\langle I \rangle} \bold A_{\bd \theta})}{\partial \bold A_{jk}} = - \alpha^{\langle I \rangle} \bold H^{jk}$ with bounded norm of $\alpha^{\langle I \rangle}$. In scenario (b), we have 
\begin{equation}
\frac{\partial (\bold I - \alpha^{\langle I \rangle} \bold A_{\bd \theta})}{\partial \bold A_{jk}} = \frac{\partial \bold A_{\bd \theta}}{\partial \bold A_{jk}} \frac{\partial (\bold I - \alpha^{\langle I \rangle} \bold A_{\bd \theta})}{\partial \bold A_{\bd \theta}}  + \frac{\partial \alpha^{\langle I \rangle}}{\partial \bold A_{jk}}  \frac{\partial (\bold I - \alpha^{\langle I \rangle} \bold A_{\bd \theta})}{\partial \alpha^{\langle I \rangle}}  = - \alpha^{\langle I \rangle} \bold H^{jk} - \bold A_{\bd \theta} \frac{\partial \alpha^{\langle I \rangle}}{\partial \bold A_{jk}}.
\end{equation}
This leads to the following,
\begin{equation}
\begin{aligned}
\bold B^{\langle I+1 \rangle}_{\bold A_{jk}} &= (\bold I - \alpha^{\langle I \rangle} \bold A_{\bd \theta}) \bold B^{\langle I \rangle}_{\bold A_{jk}} - ( \alpha^{\langle I \rangle} \bold H^{jk} + \bold A \frac{\partial \alpha^{\langle I \rangle}}{\partial \bold A_{jk}}) \bd e^{\langle I \rangle}.
\end{aligned}
\end{equation}
It remains to be shown that the norm of $\bold A \frac{\partial \alpha^{\langle I \rangle}}{\partial \bold A_{jk}}$ can be upper-bounded. Then, the Jacobian error is upper-bonded by the same order of convergence as in scenario (a),
\begin{equation}
\norm{\bold A \frac{\partial \alpha^{\langle I \rangle}}{\partial \bold A_{jk}}}_2 \leq \norm{\bold A_{\bd \theta}}_2 \norm{\frac{\partial \alpha^{\langle I \rangle}}{\partial \bold A_{jk}}}_2 = \lambda_1 \norm{\frac{\partial \alpha^{\langle I \rangle}}{\partial \bold A_{jk}}}_2.
\end{equation}
We write $\alpha^{\langle I \rangle}$ in terms of eigenvalues of $\bold A_{\bd \theta}$ (i.e. $\alpha^{\langle I \rangle} = \frac{\sum_o \zeta^{\langle I \rangle 2}_o \lambda^2_o}{\sum_b \zeta^{\langle I \rangle 2}_b \lambda^3_b}$) and take the derivative, i.e.
\begin{equation}
\frac{\partial \alpha^{\langle I \rangle}}{\partial \bold A_{jk}} = \frac{\left(\sum_o 2 \zeta^{\langle I \rangle 2}_o \lambda_o \bd v_o^{\top} \bold H^{jk} \bd v_o\right) \left(\sum_b \zeta^{\langle I \rangle 2}_b \lambda^3_b \right) - \left(\sum_o \zeta^{\langle I \rangle 2}_o \lambda^2_o\right) \left(\sum_b 3 \zeta^{\langle I \rangle 2}_b \lambda^2_b \bd v_b^{\top} \bold H^{jk} \bd v_b\right)}{(\sum_m \zeta^{\langle I \rangle 2}_m \lambda^3_m)^2}.
\end{equation}
Given the above, there exist a constant that bounds the norm of this derivative. We denote this constant by $\norm{\frac{\partial \alpha^{\langle I \rangle}}{\partial \bold A_{jk}}}_2 \leq M_{\alpha_{\text{div}}}$.

\section{Additional Experimental Details}
\subsection{Parameter Recovery for Noisy AR Models}
\subsubsection{Form of Prior Precision $\bold \Gamma_{\bd \theta}$} \label{app-ar-stat}
From \eqref{noisy-ar}, let us define $\bd z_{\leq P} := \{z_1, \ldots, z_P\}$, $\bd z_{> P} := \{z_{P+1}, \ldots, z_D\}$ and $\bd w := \{w_{P+1}, \ldots, w_D\}$.  Then, 
\begin{align} \label{cov-formula}
\begin{bmatrix}
\bold I & \bold 0 \\
\bold H & \bold L
\end{bmatrix}
\begin{bmatrix}
\bd z_{\leq P} \\
\bd z_{> P}
\end{bmatrix} = \begin{bmatrix}
\bd z_{\leq P} \\
\bd w
\end{bmatrix},
\end{align} 
where $\bold H \in \R^{(D - P) \times P}$ and $\bold L \in \R^{(D - P) \times (D-P)}$ such that
\begin{align}
\bold H := \begin{bmatrix}
- \phi_P & - \phi_{P-1} & \ldots & -\phi_1 \\
0 & - \phi_{P} & \ldots & -\phi_2 \\
0 & 0 & \ldots & \vdots  \\
0 & 0 & \ldots & -\phi_P \\
0 & 0 & \ldots & 0 \\
\vdots & \vdots & \vdots  & \vdots \\
0 & 0 & \ldots & 0 
\end{bmatrix} 
& &
\bold L := \begin{bmatrix}
1 & 0 &  0 & \ldots & 0 \\
-\phi_1 & 1 & 0 & \ldots & 0 \\
\vdots & \vdots & \vdots & \ldots & \vdots  \\
-\phi_{P-1} & -\phi_{P-2} & -\phi_{P-3} & \ldots & 0 \\
-\phi_P & -\phi_{P-1} & -\phi_{P-2} & \ldots & 0 \\
\vdots & \vdots & \vdots  & \vdots & \vdots \\
0 & 0 & 0 &  \ldots &1
\end{bmatrix}.
\end{align}
Observe that $\begin{bmatrix}
\bd z_{\leq P} \\
\bd z_{> P}
\end{bmatrix}$ is a multivariate Gaussian with mean $\bd 0$ and inverse-covariance $\bd \Gamma_{\bd \theta}$ (i.e. our object of interest).  Similarly, $\begin{bmatrix}
\bd z_{\leq P} \\
\bd w
\end{bmatrix}$ is also a multivariate Gaussian with mean $\bd 0$ and inverse-covariance $\begin{bmatrix}\bold Q_{\bd \phi}^{-1} & \bd 0 \\
\bd 0 & \sigma^{-2} \bold I \end{bmatrix}$.  Let $\bold Q_{\bd \phi}^{-1} = \sigma^{-2} \bold D$ for some matrix $\bold D$. The change-of-variables formula for probability distributions then tells us that 
\begin{align}
\bold \Gamma_{\bd \theta} = \begin{bmatrix}
\bold I & \bold 0 \\
\bold H & \bold L
\end{bmatrix}^\top 
\begin{bmatrix}\bold Q_{\bd \phi}^{-1} & \bd 0 \\
\bd 0 & \kappa^{-1} \bold I \end{bmatrix}
\begin{bmatrix}
\bold I & \bold 0 \\
\bold H & \bold L
\end{bmatrix} = \frac{1}{\kappa} \begin{bmatrix}
\bold I & \bold 0 \\
\bold H & \bold L
\end{bmatrix}^\top 
\begin{bmatrix} \bold D & \bd 0 \\
\bd 0 & \bold I \end{bmatrix}
\begin{bmatrix}
\bold I & \bold 0 \\
\bold H & \bold L
\end{bmatrix}.
\end{align}
(Note that change-of-variables also tells us that $\log \det \bd \Gamma_{\bd \theta} = \log \det \bold D  - T \log \kappa$, which we need to compute the term $\bd c_{\theta}$.)

\citet{galbraith1974inverses} show that for a \emph{stationary} AR process, we have $\bold D = \bold L_P^\top \bold L_P - \bold H_P^\top \bold H_P$, where $\bold H_P \in \R^{P \times P}$ is the first $P$ rows of $\bold H$ and $\bold L_P \in \R^{P \times P}$ is the top-left $P \times P$ block of $\bold L$. 

Next, let $\bold D$ be factorized as $\bold R \bold R^\top = \bold D$, which we can obtain through Cholesky decomposition.  This implies that $\bold \Gamma_{\bd \theta} = \bold X^\top \bold X$, where 
\begin{align}
\bold X := \frac{1}{\sqrt{\kappa}} 
\begin{bmatrix}
\bold R & \bold 0 \\
\bold H & \bold L
\end{bmatrix}.
\end{align}

We can verify that $\bold X$ is a lower triangular and banded matrix with $P + 1$ non-zero bands below the diagonal.  Similarly $\bold X^\top$ is an upper triangular and banded matrix with $P + 1$ non-zero bands above the diagonal.  It follows that $\bold \Gamma_{\bd \theta}$ has $2P + 1$ non-zero bands.  In turn, this implies that $\bold A_{\bd \theta}$ in \eqref{def-a} also has $2P + 1$ non-zero bands (since the other part of the sum is a diagonal matrix).

\subsubsection{Kalman Smoother Implementation of Exact-Gradient EM} \label{app-kalman}
Observe that one can easily write \eqref{noisy-ar} as a state-space model by defining the state
\begin{align}
\bd s_d := \begin{bmatrix} z_d \\
\vdots \\
z_{d+P - 1}
\end{bmatrix}
\end{align}
for $d = 1, \ldots, D$.  Then, the noisy AR model of \eqref{noisy-ar} is equivalent to the following state-space model:
\begin{align}
\bd s_1 &\sim \mathcal{N}(\bd 0, \bold Q_{\bd \phi}) \\
\bd s_{d} &= \bold F \bd s_{d-1} + \bd v_d, \quad \bd v_d \sim \mathcal{N}(\bd 0, \bold V) \\
y_d  &= \bd c^\top \bd s_d + w_d, \quad w_d \sim \mathcal{N}(0, W)
\end{align}
for 
\begin{align}
\bold F := \begin{bmatrix}
0 & 1 & 0 & \ldots & 0 \\
0 & 0 & 1 & \ldots & 0 \\
\vdots & \vdots & \vdots & \vdots & \vdots \\ 
0 & 0 & 0& \ldots & 1 \\
\phi_P & \phi_{P-1} & \phi_{P-2} & \ldots & \phi_1 
\end{bmatrix}, \quad 
\bold V := \begin{bmatrix}
0 & 0 & 0 & \ldots & 0 \\
0 & 0 & 0 & \ldots & 0 \\
\vdots & \vdots & \vdots & \vdots & \vdots \\
0 & 0 & 0 & \ldots & 0 \\
0 & 0 & 0 & \ldots & \kappa \\
\end{bmatrix}, \quad \bd c := \begin{bmatrix}
1 \\
0 \\
0 \\ 
\vdots \\
0
\end{bmatrix},  \quad W = \lambda.
\end{align}

The Kalman smoother can be used to obtain the distributions $p(\bd s_d | \bd y)$ for all $d$, which we can then convert into our posterior of interest $p(z_d | \bd y)$.  Kalman will require $D$ steps with complexity $\mathcal{O}(P^3)$ each because the state size is $P$.  We then follow \citet{bishop2006pattern}, Chapter 13.3.2 to compute the EM objective \eqref{qn} using the Kalman smoother outputs.

\subsubsection{Experimental Settings} \label{ar-exp-settings}
It is not straightforward to directly optimize $\bd \phi$ over the space of \emph{stationary} noisy AR processes.  Thus, we parameterize $\bd \phi := f(\bd \gamma)$, where $\bd \gamma \in [-1, 1]^P$ are partial auto-correlations and $f$ is the transformation defined by \citet{barndorff1973parametrization}.  Gradient descent is performed over $\bd \gamma, \log \kappa$ and $\log \lambda$.   For each algorithm (i.e. gradient EM, probabilistic unrolling), we perform 200 iterations of gradient descent with the Adam optimizer and learning rate $0.1$.

Each ground-truth value $ \gamma^\star_p$ is randomly initialized between $[-1, 1]$.  Similarly, $\kappa^\star$ and $\lambda^\star$ are randomly initialized between $[0.1, 10]$ (in log-space).  

\subsubsection{Comparison with VAEs} \label{ar-vae-comp}
To provide another baseline of comparison for probabilistic unrolling, we fit the model in \eqref{noisy-ar} using a variational auto-encoder (VAE) \citep{kingma2013auto}.  The decoder of the VAE is the generative model in \eqref{noisy-ar}, and the encoder of the VAE is a black-box deep neural network.  To tune the VAE, we search over different architectures (i.e. 1, 2, 3, 4 layer models), different activations (i.e. ReLU, Sigmoid, Tanh, LeakyReLU) and adjust weights for the different parts of the VAE loss (i.e. the weight $\beta$ in $\beta$-VAE \citep{higgins2017beta}).

However, we find that VAEs perform poorly in parameter recovery for the noisy AR model.  In Table \ref{sample-table}, we observe that probabilistic unrolling (like EM) estimates all the true parameters $\{\phi^\star, \kappa^\star, \lambda^\star\}$ to within 1\% error for $D = 30{,}000$ time points.  On the other hand, the (optimally-tuned) VAE obtains $41.5 \pm 14.1\%$ error for $\phi$, $215.93 \pm 223.5\%$ error for $\kappa$, and $37.9 \pm 23.75\%$ error for $\lambda$.  We hypothesize that this poor performance is due to the VAE's biased objective in comparison to EM (and PU), which suffers from mean-field's inability to model covariance in the latent posterior; this is especially detrimental for this problem, because there is a lot of rich covariance structure across time.

\subsection{Bayesian Compressed Sensing of Sparse Signals}

\subsubsection{Woodbury Matrix Identity for Exact-Gradient EM} \label{app-woodbury}
The Woodbury matrix identity is a property from linear algebra that allows us to compute $\bold \Sigma_{\bd \theta}$ for Bayesian compressed sensing by inverting a $M \times M$ matrix instead of a $D \times D$ one.  Since typically $M < D$ in compressed sensing, this can lead to computational savings in time from a practical standpoint (but perhaps not an asymptotic one, because $M$ often needs to grow linearly with $D$ in compressed sensing applications).  Note that it does not lead to computational savings in space, because the full matrix $\bold \Sigma_{\bd \theta}$ is still computed in the end.  

Instead of computing $\bold \Sigma_{\bd \theta}$ through \eqref{mvn}, we equivalently have
\begin{align}
\bold \Sigma_{\bd \theta} = \bold \Gamma_{\bd \theta}^{-1} - \bold \Gamma_{\bd \theta}^{-1} \bold \Phi_{\bd \theta}^\top \bold \Omega^\top (\bold \Omega \bold \Psi_{\bd \theta}^{-1} \bold \Omega^\top + \bold \Omega \bold \Phi_{\bd \theta} \bold \Gamma_{\bd \theta}^{-1} \bold \Phi_{\bd \theta}^\top \bold \Omega^\top)^{-1} \bold \Omega \bold \Phi_{\bd \theta} \bold \Gamma_{\bd \theta}^{-1}.
\end{align}
Note that for Bayesian compressed sensing, both $\bold \Gamma_{\bd \theta}^{-1}$ and $\bold \Psi_{\bd \theta}^{-1}$ can be cheaply computed, because these are both diagonal matrices. 

\subsubsection{Experimental Settings} \label{app-bcs-details}
The NIST dataset is accessed at \url{https://www.nist.gov/srd/nist-special-database-19}.

We scale all raw image pixels in NIST from $[0, 255]$ to the range $[0, 1]$.  We add independent, pixel-wise Gaussian noise to the undersampled 2D Fourier transform with standard deviation $\sigma = 0.01$.  During model fitting, we use the Adam optimizer with learning rate 1.0.  We optimize the parameters $\bd \alpha, \beta$ in log-space.  Each component of $\log \bd \alpha$ is initialized as randomly drawn from $\mathcal{N}(0, 1)$.  The value $\log \beta$ is initialized as 0.  

For probabilistic unrolling, we use the preconditioned conjugate gradient algorithm.   The preconditioner we employ is the diagonal preconditioner $\bold M$ introduced in \citet{lin2022covariance} for sparse Bayesian learning; $\bold M^{-1}$ is a diagonal matrix with diagonal $\bd m$, where 
\begin{align}
m_j := \frac{1}{\alpha_j + \beta}.
\end{align}

\subsubsection{Sample Images}
\begin{figure}[ht]
\vskip 0.2in
\begin{center}
\centerline{\includegraphics[scale=0.4]{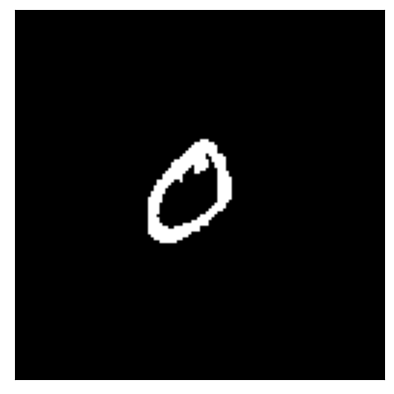} \includegraphics[scale=0.4]{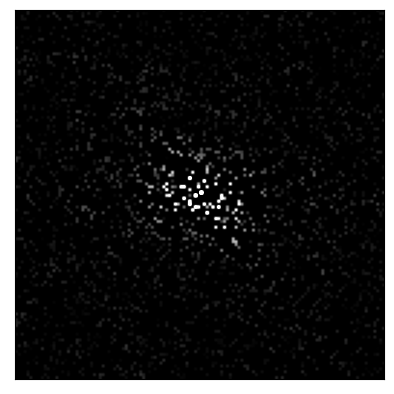} \includegraphics[scale=0.4]{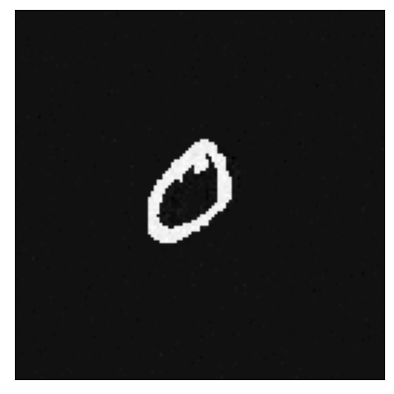}}
\caption{From left to right: the true signal, the Fourier measurement, and the probabilistic unrolling reconstruction.}
\label{samp-img}
\end{center}
\vskip -0.2in
\end{figure}

In Figure \ref{samp-img}, we provide sample images of the true signal $\bidx {\tilde z} n$, its 15\% undersampled and noisy Fourier transform measurement $\bidx \ty n$, and a reconstruction provided by probabilistic unrolling $\bidx \mu n$.

\subsubsection{Results Breakdown by Digit Type} \label{app:breakdown}
Table \ref{tab:breakdown} presents a breakdown of the compressed sensing results by digit type.  The average is given in Table \ref{bcs-results}.
\begin{table}[!h!]
\vskip 0.15in
\begin{center}
\begin{small}
\caption{Compressed sensing results broken down by digit type.} \label{tab:breakdown}
\begin{tabular}{lcccc}
\toprule
Digit Type & $r(\bd \mu^\text{EM}, \btilde z)$ & $r(\bd \mu^\text{PU}, \btilde z)$ & EM Time & PU Time  \\
\midrule 
0 & 3.89\% & 4.01 \% & 1475 s & 21 s\\
1 & 4.76\% & 4.31 \% & 1483 s & 21 s\\
2 & 4.61\% & 4.43 \%  & 1473 s & 21 s\\
3 & 3.96\% & 3.90 \%  & 1508 s & 21 s\\
4 & 7.20\% & 8.56 \%  & 1485 s & 21 s\\
5 & 5.22\% & 4.88 \%  & 1494 s & 21 s\\
6 & 4.17\% & 3.94 \% & 1510 s & 21 s\\
7 & 4.67\% & 4.59 \%  & 1467 s & 21 s\\
8 & 4.27\% & 4.08 \%  & 1468 s & 21 s\\
9 & 4.83\% & 4.46 \%  & 1448 s & 21 s \\
\bottomrule
\end{tabular}
\end{small}
\end{center}
\vskip -0.1in
\end{table}

\begin{figure}[h!]
\vskip 0.2in
\begin{center}
\centerline{\includegraphics[scale=0.55]{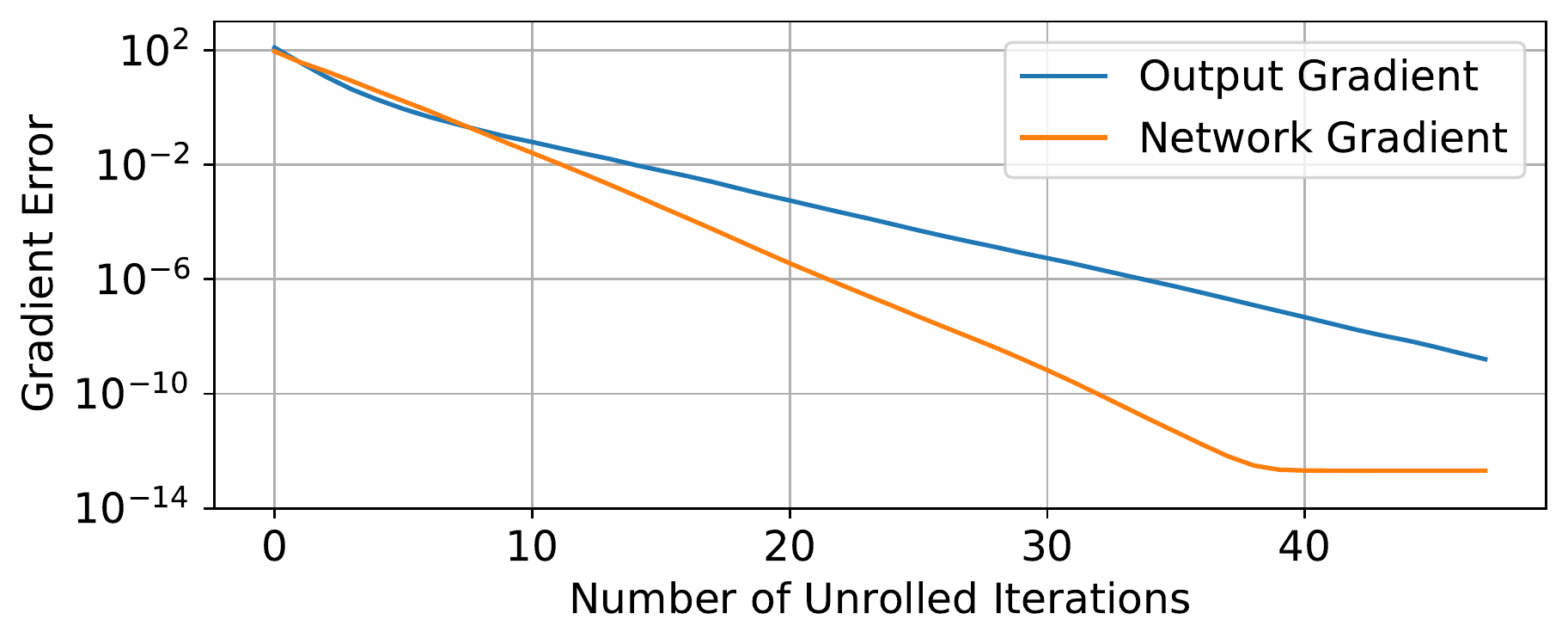}}
\caption{Gradient convergence as a function of iterations $I$.}
\label{grad-converge}
\end{center}
\vskip -0.2in
\end{figure}

\subsubsection{Super-Efficiency of Network Gradient}
In Figure \ref{grad-converge}, we empirically show that the network gradient \eqref{net-grad} converges faster to the Monte Carlo gradient \eqref{mc-grad} than the output gradient \eqref{out-grad} for our preconditioned conjugate gradient solver.

\subsection{Collaborative Filtering through Factor Analysis}

\subsubsection{Experimental Settings} \label{app:fa}
The MovieLens datasets are accessed at \url{https://grouplens.org/datasets/movielens/}.

We follow the general experimental setup of \citet{sedhain2015autorec}.  For each dataset, we perform a 90\%-10\% train-test split of the ratings data.  We then set aside 10\% of the training set as a validation set.  Both models (gradient EM and probabilistic unrolling) are trained with the Adam optimizer, learning rate 0.001, and a user mini-batch size of 25 (with gradient accumulation over four mini-batches for an overall batch size of 100).  For the ML-1M dataset, we train the model for 20 epochs and evaluate the model on the validation set every 500 gradient steps.  For the ML-10 M and ML-25 M datasets, we train the model for 5 epochs and evaluate the model on the validation set every 2{,}000 gradient steps.  The checkpoint with the best validation set RMSE is used for final evaluation on the test set.  

During training, we only train on users with at least one rating in the training set.  Thus, there may be users in the validation/test sets that do not appear during training.  Following \citet{sedhain2015autorec}, we always predict a rating of 3 for these users.  All other model rating predictions are clipped to be in the range [1.0, 5.0] before evaluation of RMSE. 

\subsubsection{Comparison with VAEs} \label{fa-vae-comp}
Similar to Appendix \ref{ar-vae-comp}, we report results for a VAE baseline on the collaborative filtering task.  Searching over architectural options, we find that the best-performing architecture had a two-layer encoder with $3{,}000$ units in the hidden layer and ReLU activations.  However, even with our extensive tuning, we observe that VAEs generally perform worse and have higher computational costs in comparison to probabilistic unrolling. The VAE obtains 0.8849 RMSE (compared to 0.8436 for PU) on MovieLens 1-m and 0.8366 RMSE (compared to 0.7796 for PU) on MovieLens 10-m. The VAE also has approximately 1.5x the time cost and 2x the memory cost of PU (due to the separate inference encoder network).

\end{document}